# FDNet: Focal Decomposed Network for Efficient, Robust and Practical Time Series Forecasting

Li Shen[a], Yuning Wei[a,*], Yangzhu Wang[a] and Huaxin Qiu[a]

[a] *Beihang University,*
Postal Address: RM.807, 8th Dormitory, Dayuncun Residential Quarter, No.29, Zhichun Road Beijing 100191 P.R. China

**Abstract:** This paper presents FDNet: a *F*ocal *D*ecomposed *Net*work for efficient, robust and practical time series forecasting. We break away from conventional deep time series forecasting formulas which obtain prediction results from universal feature maps of input sequences. In contrary, FDNet neglects universal correlations of input elements and only extracts fine-grained local features from input sequence. We show that: (1) Deep time series forecasting with only fine-grained local feature maps of input sequence is feasible upon theoretical basis. (2) By abandoning global coarse-grained feature maps, FDNet overcomes distribution shift problem caused by changing dynamics of time series which is common in real-world applications. (3) FDNet is not dependent on any inductive bias of time series except basic auto-regression, making it general and practical. Moreover, we propose focal input sequence decomposition method which decomposes input sequence in a focal manner for efficient and robust forecasting when facing Long Sequence Time series Input (LSTI) problem. FDNet achieves competitive forecasting performances on six real-world benchmarks and reduces prediction MSE by 38.4% on average compared with other thirteen SOTA baselines. The source code is available at https://github.com/OrigamiSL/FDNet.

## 1. Introduction

Deep time series forecasting develops rapidly in recent years owing to more pressing demands [1-6] of handling complicated non-stationary time series [7, 8]. At present, there exist deep time series forecasting networks in diverse formulas, including networks based on RNN [9, 10] /CNN [11, 12] /Transformer [13-15], networks based on end-to-end [12, 16, 17] /self-supervised [18, 19] forecasting format, etc. However, they all obey similar forecasting procedures which can be simply divided into 3 steps as shown in Figure 1: (a) Embedding input sequence into latent space; (b) Feature extraction of input sequence; (c) Project input sequence latent representation into prediction sequence. Within the second step, nearly all methods have mechanisms to extract universal correlation information of different input elements to seek universal/global features of input sequences such as attention mechanism [20-22], dilated convolution [11, 12, 18], etc. We call them *I*nput *C*orrelation *O*riented *M*echanism(s) and ICOM for short. We will expain ICOM more detailedly in Section 4. However, time series forecasting task is intended to pursue **connections of previous and future sequence** instead of only concerning **the correlation information or universal features of previous sequences**. So here comes the question: *Does ICOM necessary for time series forecasting?* We analyze this question from three perspectives in Section 4 and show that network without ICOM is still capable of doing time series forecasting and even can do better. Therefore, we propose FDNet, a *F*ocal *D*ecomposed time series forecasting network. FDNet uses *decomposed forecasting formula* and its differences with existing rolling [8, 12, 20] and direct [9-11] forecasting formulas are illustrated in Figure 2. Built upon direct forecasting formula where forecasting processes of prediction elements are decomposed, decomposed forecasting formula further decomposes feature extraction processes of input elements. Hence, FDNet is composed of basic linear projection layers to extract local fine-grained feature maps of input sequence to get rid of conventional ICOMs and canonical convolutions to stabilize feature extraction processes when handling outliers of input sequences.

Apart from the necessity of ICOM, currently there also exists another problem which is often ignored, i.e., the Long Sequence Time series Input (LSTI) problem [23-25]. Though it is believed that networks which are able to extract long-term dependencies, e.g., Time Series Forecasting Transformers (TSFTs) [8, 17, 22], have already gotten rid of LSTI problem, [26] points out that even TSFTs will suffer performance drop if excessively prolonging input sequences over a certain borderline as the problem of overfitting will overwhelm benefits of obtaining long-term dependency. Obviously, a qualified forecasting network which can capture potential long-term dependency shall at least not suffer performance drop if prolonging input sequence. Moreover, it is unacceptably time-consuming and expensive if input sequences are too long for most of forecasting networks, especially TSFTs [13, 20]. Therefore, a novel input sequence decomposition strategy which not only can deal with LSTI problem but also can limit parameter explosion with the prolonging of input sequence is needed. Motivated from the discovery of [26] that later input sequence elements are more related to prediction sequences and Focal Transformer [27], we propose *focal input sequence decomposition method* to help networks deal with LSTI problem. Focal input sequence decomposition divides input sequence into several consecutive sub-sequences in a focal manner according to their temporal distances with prediction elements. Closer a sub-sequence is to prediction elements, shorter it is and more feature extraction layers it has. As a result, connections of input and prediction elements will become weaker and shallower as their temporal distances get farther. Moreover, with the prolonging of input sequence, extra parameters of extra input sequence will also not suffer parameter explosion in that they own fewer feature extraction layers. Focal input sequence decomposition renders FDNet not only accurate but also efficient, especially when handling LSTI problems.

---

* Corresponding author
*E-mail address:* yuning@buaa.edu.cn (Y. Wei)

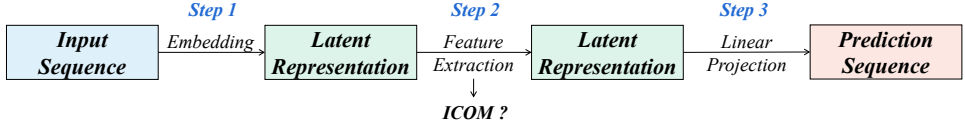

**Figure 1:** An overview of the similar forecasting procedure of all deep time series forecasting models. It contains three steps and ICOM is employed in the second step if needed.

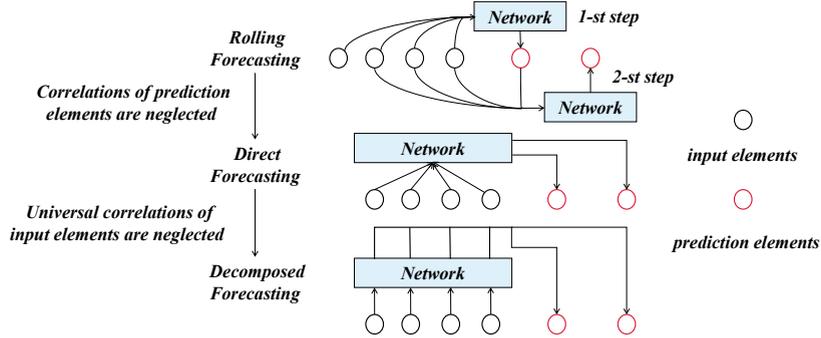

**Figure 2:** Connections and differences among three forecasting formulas. Correlations of prediction elements are neglected by direct forecasting formula while universal correlations of input elements are further neglected by our proposed decomposed forecasting formula.

Our main contributions are summarized as below:

1. We propose a novel decomposed forecasting formula. Built upon direct forecasting formula, it decomposes both forecasting processes of prediction elements and feature extraction processes of input elements.
2. We propose FDNet which uses decomposed forecasting formula. It is only composed of basic linear projection layer and CNN, thus its architecture is very simple. However, FDNet is accurate, efficient and robust in time series forecasting.
3. We propose focal input sequence decomposition method to deal with long-standing LSTI problem. It gives forecasting networks capability of handling extremely long input sequences without suffering overfitting problem, performance drop and parameter explosion.
4. Extensive experiments over six benchmark datasets show that FDNet outperforms thirteen SOTA forecasting methods by 36.2%/40.5% for multivariate/univariate forecasting on average.
5. Ablation study of focal input sequence decomposition method demonstrates that it is competitive in dealing with LSTI problems. Moreover, it is general enough to couple with not only decomposed forecasting formula but also other forecasting formulas owning ICOM.

## 2. Meanings of Abbreviations and Phrases

Meanings of mentioned abbreviations and phrases used in this paper are shown in Table 1.

**Table 1**
Meanings of abbreviations and phrases

| Abbr./Phrase | Meaning |
|---|---|
| LSTI | *L*ong *S*equence *T*ime series *I*nput |
| TSFT | *T*ime *S*eries *F*orecasting *T*ransformer |
| ICOM | *I*nput *C*orrelation *O*riented *M*echanism |
| RR | *R*eject *R*ate |
| MSE | *M*ean *S*quare *E*rror |
| MAE | *M*ean *A*bsolute *E*rror |
| SMAPE | *S*ymmetric *M*ean *A*bsolute *P*ercentage *E*rror |
| $L_{in}$ | Length of input sequence |
| $L_{out}$ | Length of output sequence |
| $V$ | Number of variates |
| $D$ | Dimension of embedding |
| Std | *St*andard *d*eviation |
| GPU | *GPU* memory occupation |
| ATPE | *A*verage *T*raining time *P*er *E*poch |
| TIT | *T*otal *I*nference *T*ime |
| TLP | *T*otal *L*earnable *P*arameters |

## 3. Related Works

### 3.1. Deep Time Series Forecasting

Time series forecasting has wide application in various domains: e.g., electricity prediction [28], traffic forecasting [29-31], sensor-based recognition [32] and COVID-19 pandemic analysis [5, 6, 33]. In order to handle long-term and complicated time series, machine learning techniques including neural networks are applied to time series forecasting gradually dominant the research direction. Time series forecasting networks are normally categorized by their network architectures (RNN [9, 10], CNN [11, 12], Transformer [8, 15, 20, 21], GNN [34, 35], etc.) or learning algorithm (supervised end-to-end [14, 22] or self-supervised representation learning [18, 19]). However, we categorize them by three different criteria as shown below to discuss deep time series forecasting from other perspectives.

### 3.2. Rolling/Direct Forecasting Formula

It is believed that time series forecasting starts from rolling forecasting. Traditional models like ARIMA [36-38], State Space Estimation [39], ES [40] are all rolling forecasting models. Given input sequence $\{z_{i,1:t_0}\}_{i=1}^{V}$, the prediction sequence $\{z_{i,t_0+1:T}\}_{i=1}^{V}$ is obtained by rolling forecasting strategy which predicts the prediction sequence through $(T - t_0)$ steps. However, in virtue of the end-to-end property of neural network, deep time series forecasting network has another option, i.e., unrolled forecasting formula [12, 20, 21]. Given input sequence $\{z_{i,1:t_0}\}_{i=1}^{V}$, the prediction sequence $\{z_{i,t_0+1:T}\}_{i=1}^{V}$ is obtained by one-forward strategy which predicts the entire prediction sequence in one step. Hence, rolling forecasting is also called recursive forecasting while unrolled forecasting is also called direct forecasting [41]. Comparing these two forecasting formulas, it can be deduced that direct forecasting formula is more efficient in that it only needs one forward propagation process to obtain the entire prediction sequence. Moreover, inputs of direct forecasting models are all known and definite while inputs of rolling forecasting models are partially unknown and inferred from known inputs and models themselves. It is obvious that direct forecasting models theoretically suffer slighter error accumulation so that direct forecasting formula is more preferred and common in recent researches [9, 12, 22]. Here we make specific explanations for direct Time Series Forecasting Transformer (TSFT) [13, 20, 21] as they own masked self-attention mechanisms which seems like inferring prediction elements from themselves and contradicts to our saying that direct forecasting formula makes inference processes of different prediction elements decomposed and independent from each other. However, notice that inputs of their mask self-attention mechanisms are either zero-padded units [13, 20] or decomposed parts of input sequences [8, 22]. Neither of these inputs can reflect essential relationships insider prediction sequences. Some state-of-the-art TSFTs [8, 22] also notice this phenomenon, thus they do not employ masked self-attention mechanisms within their networks but still achieve promising forecasting performances.

### 3.3. Variable-agnostic/-specific Forecasting Formula

Multivariate forecasting starts to be practical after the involvement of machine learning. It has two multivariate oriented forecasting formulas categorized by [17], i.e., Variable-agnostic and Variable-specific forecasting formula.

Models with variable-agnostic forecasting formula [20, 21, 42] have the same projection matrices for all variables while models with variable-specific forecasting formula [43-45] have distinct (decomposed) projection matrices for them. In other words, variable-agnostic formula assumes that variables are closely related to each other and have the same statistical properties and variable-specific formula assumes the opposite. RTNet [26] also points out that variable-agnostic formula only works if we have priori knowledge that variables are closely relevant throughout the whole time span otherwise this formula will make models suffer heavy over-fitting problem. Consequently, experiment results of variable-specific models [17] show that variable-specific formula behaves better in most of benchmark datasets/real-world applications.

### 3.4. Compound/Decomposed Forecasting Formula

Time series forecasting networks can also be categorized by their input sequence feature extraction methods. Those methods which own ICOM belong to compound forecasting methods. Traditional forecasting methods [36, 37, 39] and most of deep time series forecasting networks are compound forecasting methods. For instance, classical RNN forecasting networks including DeepAR [10], LSTNet [9], CNN forecasting networks including TCN [11], TS2Vec [18] and TSFTs including Informer [20], LogTrans [13] are all compound forecasting methods. They may own hierarchical feature extraction mechanisms or pyramid networks which only extract feature maps from partial input sequence, i.e., ICOMs. However, they all have partial networks extracting universal feature maps from the whole input sequence. Recently, some researches point out that decomposing input sequence into season and trend [22] or even season, level and growth [8] gives networks more competitive forecasting performances. Obviously, these researches propose their methods based on inductive bias that time series can be decomposed like this, which means that their decompositions are beneficial under limited occasions. At least, non-stationary time series which are more common in real-world applications cannot easily be described like this so that these methods are not very practical. We have analyzed above statement in Section 4. Moreover, their decompositions do not get rid of universal coarse-grained feature extraction. Though season and trend terms are decomposed from time series, they are still universal representations of specific properties of time series, i.e., these kinds of networks still own ICOM and belong to compound forecasting methods. In fact, there exist few decomposed forecasting networks [16, 46, 47] which are similar to our concepts. However, their methods in practice are quite different from ours. They abandon embedding layers and process input sequence not in latent space but with only linear projection layers. They also achieve the goal of decomposed forecasting, yet they contradict concepts of deep learning. Such shallow representations of input sequence will limit expression skills and potentials of forecasting networks. As a result, a practical decomposition implementation form not limited in season-trend decompositions and shallow feature extraction is needed, which is the motivation of this paper.

FDNet attempts to take the primary advantages of all above methods into account and overcomes their limitations. It employs decomposed forecasting strategy to replace rolling/direct forecasting strategy so that universal correlations within input sequence and between input and prediction sequences are both neglected. Together with focal input sequence decomposition method, FDNet can handle LSTI problem efficiently, effectively and robustly. It applies specific 2D convolutions and linear projection layers to separate features of sequence from different variates while making them share the same weight matrices. Therefore, the fact that different variates own distinct properties is respected and the number of model parameters is also controlled.

## 4. Necessity Analysis of ICOM

Above all, we first provide more detailed definition and explanations for ICOM and then discuss whether a time series forecasting model is still established if removing its ICOM from three perspectives to demonstrate the rationality of decomposed forecasting formula.

### 4.1. Definition and Explanations of ICOM

ICOM for time series forecasting models is a mechanism to extract universal feature maps of input sequences. In practice, ICOM aims to seek correlations of each input element and other input elements through diverse formats. Self-attention [13, 14, 17, 20] and dilated convolution [11, 18, 19] are typical ICOMs. Attention scores computed in self-attention represent correlations of all input elements while dilated convolution ensures that each input element can build correlations with others via expanding receptive field exponentially. Except these basic ICOM architectures, some models [8, 21, 22] additionally extract universal feature maps in frequency domain by employing signal processing methods. In essence, ICOM is an artifact of imposing some inductive biases to properties of input sequence. It could enhance network forecasting performance only when these assumed universal correlations are identical for all windows throughout the entire dataset.

## 4.2. Time Series Forecasting Definition

Time series forecasting is defined as the task of predicting future time series through current and previous time series. Given input sequence $\{z_{i,1:t_0}\}_{i=1}^{V}$, the task is to obtain the prediction sequence $\{z_{i,t_0+1:T}\}_{i=1}^{V}$. $V$ is the number of variates; $t_0$ denotes the length of input sequence; $(T - t_0)$ refers to the length of prediction sequence. It can be observed that a time series forecasting model without ICOM is still able to do the forecasting task in that only the projection of input sequence to prediction sequence, i.e., step 3 in Figure 1, is prerequisite [36, 37, 48]. In other words, ICOM is only a feature extraction technique of input sequence instead of a necessary component of a forecasting network though it is widely used.

## 4.3. Network Expression Skills

We compare expression skills of forecasting networks with/without ICOM to check effects of ICOM. As a downstream task of time series analysis, time series forecasting focuses on finding the correlation of input and prediction sequences, as mentioned in the former perspective. As a result, expression skill here denotes the expression skill of prediction sequences by input sequences. Expression skills of networks with/without ICOM have different dominant domains in time series forecasting which are determined by whether universal feature map of input sequence exists or is beneficial for forecasting. However, the existence of universal feature map relies on the inductive bias or manual assumption of time series properties, especially those networks assume the season-trend decomposition of time series [8, 21, 22]. Networks with ICOM will leverage from these manual assumptions if time series dealt with really have supposed properties, otherwise they will suffer from severe forecasting performance turbulence and over-fitting problem. We choose Autoformer [21] and Autoformer vs. ETSformer [8] as examples to verify aforementioned statements respectively by evaluating their forecasting capabilities. Prediction lengths are within {96, 192, 336, 720}. Firstly, Autoformer is partially known for its feature extraction capability of universal correlations of input sequences (Auto-Correlation mechanism). It could be observed from Figure 3(a) that MSEs of Autoformer own long errorbars, i.e., large standard deviations, under univariate forecasting of ETTm$_2$, which means that Autoformer suffers from severe forecasting performance turbulence. Moreover, following season-trend decomposition of Autoformer, ETSformer additionally decomposes time series into season, level and growth to enhance its capability of extracting universal features in more specific formats. However, as Figure 3(b)/(c) show, ETSformer surpasses Autoformer under ETTh$_1$ but performs much worse than Autoformer under ECL during univariate forecasting. It illustrates that more complex inductive bias employed by ETSformer to universal feature extraction of input sequences leads to overfitting problem under infeasible conditions. On contrary, networks without ICOM do not leverage from any assumptions of time series properties except basic auto-regression, making it less specific but more general and robust, especially when dealing with real-world time series shown as below.

## 4.4 Real-world Time Series

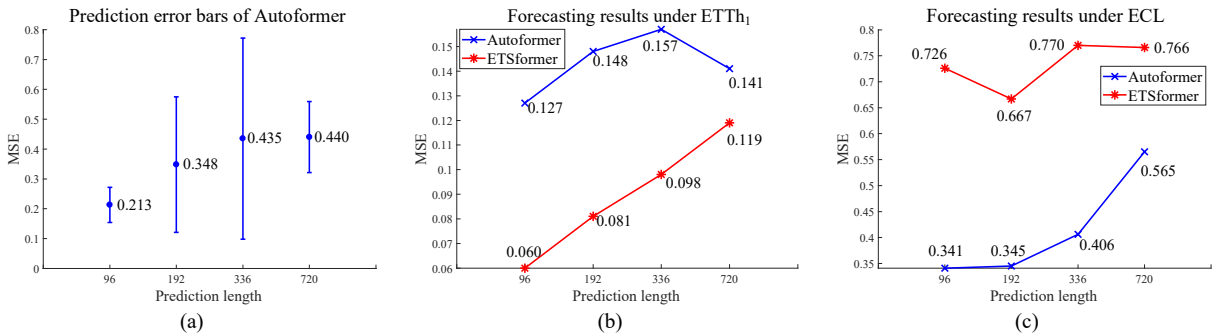

**Figure 3:** (a) Visualization of MSEs with errorbars of Autoformer during univariate forecasting under ETTm$_2$. Each blue line denotes the errorbar of MSE with different prediction lengths and their middle blue dots with numerical values represent their average MSEs respectively. (b) and (c) present forecasting MSEs of Autoformer/ETSformer during univariate forecasting under ETTh$_1$ and ECL.

**Table 2**
Results of KS test

| Dataset | ETTh$_1$ | | | ETTm$_2$ | | | ECL | | | Traffic | | | Exchange | | | Weather | | |
|---|---|---|---|---|---|---|---|---|---|---|---|---|---|---|---|---|---|---|
| Metrics | RR | Mean | Std | RR | Mean | Std | RR | Mean | Std | RR | Mean | Std | RR | Mean | Std | RR | Mean | Std |
| Values | 98.2% | 0.012 | 0.103 | 98.4% | 0.009 | 0.089 | 66.4% | 0.108 | 0.221 | 92.2% | 0.031 | 0.118 | 95.3% | 0.018 | 0.097 | 86.0% | 0.045 | 0.164 |

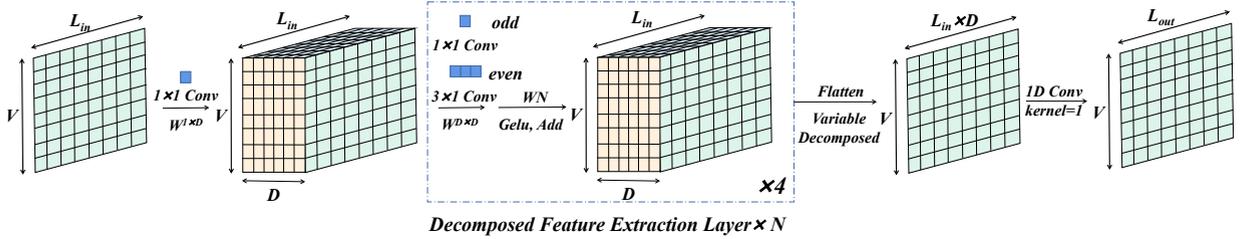

**Figure 4:** An overview of the architecture of FDNet. It decomposes feature extraction processes of different input elements and different variates. Its main components are $N$ decomposed feature extractor layers (blue trapezoid), each containing four 2D convolutional layers. Weight Normalization [51], Gelu activation [52] and res-connection [53] are combined with each 2D convolutional layer. $L_{in}$: the length of input sequence; $L_{out}$: the length of prediction sequence; $V$: the number of variables; $D$: the dimension of embedding.

Finally, we show that networks without ICOM are more practical in dealing with real-world time series. Recent researches [7, 8] have discovered that non-stationary time series, which most of real-world time series are, have different statistical properties or dynamics for local windows spanning different time stamps. This phenomenon is described as distribution shift problem [7]. We also verify it here by Kolmogorov-Smirnov Test [49, 50] on target variates of six real-world benchmark datasets. More details of this experiment are shown in Section 6.11. We randomly select 1000 sub-sequences of length 96 for each dataset and separately calculate Kolmogorov-Smirnov statistics, i.e., P-values, of the first selected sub-sequence and the rest. Results are shown in Table 2. Using a 0.05 P-value as margin statistics, it could be observed that all six datasets have extremely different local dynamics as reject rates (RRs) are all very high and standard deviations of P-values are also very big compared with the margin value 0.05. This means that universal representations for all local windows are formidable or even impossible to extract for real-world time series. Therefore, networks with ICOM are easier to get stuck at local optimum, thus their prediction performances are easily affected by random weight initialization. When statistical properties or dynamics they extract are suitable for most of local windows, their forecasting performances will be better in general, otherwise their forecasting performances will be worse and unstable. This also explains unsatisfactory prediction results of Autoformer/ETSformer shown in the former sub-section. Though some advanced methods have been proposed to alleviate this problem [7, 17], they cannot completely solve this problem in that they do not change the forecasting formula essentially. However, networks without ICOM do not have such problems as they abandon the process of extracting global representations of input sequence, which solves this problem from the source.

From above three perspectives of analysis, it can be inferred that networks without ICOM are feasible, general, robust and practical so that decomposed forecasting formula is rational.

## 5. Model Architecture

### 5.1. FDNet without ICOM

The architecture of FDNet with decomposed forecasting formula is mainly composed of linear projection layers. An overview of it is shown in Figure 4. The whole network has $(N + 2)$ layers where the first layer is the embedding layer, the last layer is the projection layer and the rest are $N$ decomposed feature extraction layers. Each decomposed feature extraction layer contains four convolutional layers. Detailed components of decomposed feature extraction layer are shown in Figure 5. *Odd layers are $1 \times 1$ convolutional layers which has the same function with Perceptron.* Except the last projection layer, we use 2D convolution, where two dimensions respectively correspond to temporal dimension and variate dimension, instead of [19] commonly used 1D convolution. *The kernel size of the variate (variable) dimension will always be one to make sure that element values of different variates will not influence each*

*other and meet the need of decomposed forecasting formula that correlations of input elements are neglected.* This replacement is motivated by variable-specific forecasting methods [17, 43, 45] which treat sequences from different variables as different instances. They point out that sequences of different variates will have different properties with each other in reality. However, employing different projection matrices for different variables is very expensive for forecasting conditions with hundreds of variates. FDNet does not mix values of different variates but gives them the same weight matrices through specific 2D convolution. This is *a balance of* variable-specific methods which completely splits different variates and the opposite variable-agnostic methods [20, 21, 42] which completely mixes different variates during forecasting. *Even layers are $3 \times 1$ convolutional layers* which is a little contradictory to the concept of decomposed forecasting formula. However, pure element-wise feature extraction of input elements will make model susceptible to outliers [54, 55]. Convolutional layers used here only have stride of 1 to *enhance the locality and smooth anomalies*. It is *a tradeoff design* between decomposed forecasting formula and realistic forecasting situations with considerable outliers. Besides, several convolutions will only make receptive fields of input elements contain few adjacent elements, which still ensures the local fine-grained feature extraction of input sequences. Consequently, feature extractions of elements from different variables and time stamps are all decomposed in FDNet. Finally, extracted feature maps of different variates are separately flattened and a 1D convolutional layer whose kernel size is 1 is used to obtain separate prediction results for separate variates. It could be seen that prolonging prediction length will only result in more parameters of this 1D convolutional layer on account of this final output design. Thus, unlike many other methods, FDNet will not suffer non-negligible parameter explosion with the increase of forecasting horizon. We will also empirically prove this statement in the following experiments (Table 11 and 12).

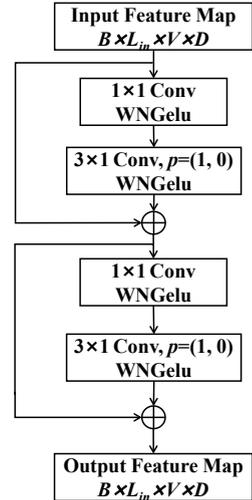

**Figure 5:** The architecture of decomposed feature extraction layer. $B$: batch size; $p$: padding.

### 5.2. Focal Input Sequence Decomposition

How focal input sequence decomposition works with forecasting networks is depicted in Figure 6. The latest sub-sequence of input sequence has the shortest length but has the most feature extraction layers. When it goes to farther regions, decomposed sub-sequence gets longer and feature map extracted from it gets shallower. Proportions of input sequence comprised by different sub-sequences approximately form a geometric series with common ratio of $\alpha$. We set $\alpha$ to 0.5 throughout following sections and experiments. In fact, $\alpha$ could be set to other positive numbers which are smaller than one, however these make no difference to the essence of focal decomposition. The number of input sub-sequences divided by focal input sequence decomposition method is a hyper-parameter denoted by $f$. For instance, if input sequence is consecutively splitted into 4 parts by focal decompostion method like Figure 6, then proportions will be {1/2, 1/4, 1/8, 1/8}. The latest sub-sequence takes the proportion of 1/8 instead of 1/16 in order to make the sum of proportions be 1. Furthermore, feature extractions of different sub-sequences and projections of them to output prediction sequence are all mutually independent. As a result, focal input sequence decomposition method effectively allocates complexity levels to different input sub-sequence independently according to their temporal distances with prediction sequence. Networks with focal input sequence decomposition method is now able to deal with LSTI problem without gaining considerable parameters and suffering performance drop with prolonging the input sequence length. When combining FDNet with focal input sequence decomposition, decomposed feature extraction layers in Figure 4 will take formats in Figure 6.

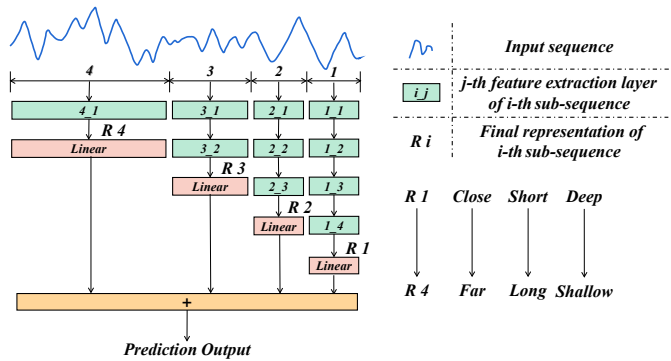

**Figure 6:** The architecture of focal input sequence decomposition. Final representations of different sub-sequences are from temporally close to far; short to long; deep to shallow.

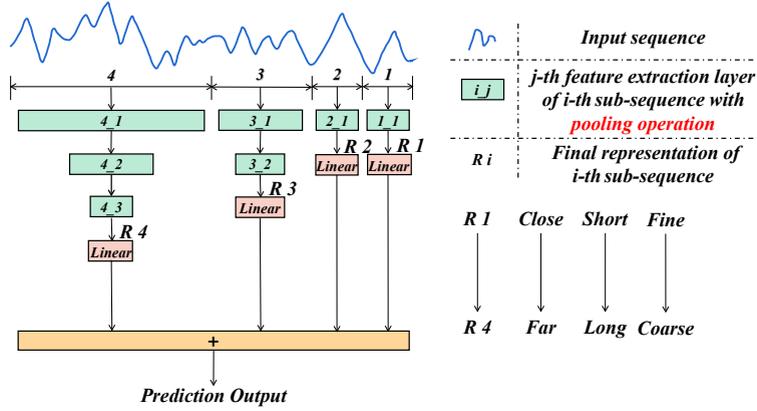

**Figure 7:** An application of focal input sequence decomposition method combined with ICOMs. Its feature extraction layers additionally contain pooling operations. Final representations of different sub-sequences are from temporally close to far; short to long; fine to coarse.

### 5.3. FDNet vs other TCNs

As components of FDNet contain convolutional layers, it is necessary to present differences of FDNets with other Temporal Convolution Networks (TCNs) [10, 12, 18]. The biggest one is that their usages of convolutions are different. Previous works focus on modifying convolutional layers or combining them with different other techniques such as representation learning [18, 19], binary tree [12], etc, to obtain universal/global feature maps of input sequence. The most well-known and widely used modification is causal dilated convolution [11, 18, 19]. In contrary, FDNet only uses the basic function of convolution, i.e., enhancing locality of neural networks. Determined by its decomposed forecasting formula, FDNet does not extract global/universal feature maps which is another big difference of it with other TCNs.

### 5.4. Focal with ICOMs

Though focal input sequence decomposition method absolutely tallies with decomposed forecasting formula, it does not mean that it cannot be applied to other forecasting formulas owning ICOM. When combining forecasting formulas with ICOMs and focal input sequence decomposition, ICOMs can be applied within each sub-sequence to extract universal features of sub-sequences. We provide a possible application which is shown in Figure 7 and call this *F*ocal *U*niversal *Net*work (FUNet). Slightly different from the application of focal input sequence decomposition in decomposed forecasting formula, farther sub-sequences have more feature extraction layers. However, the core idea is invariant in that feature extraction layers here have additional pooling operations[1] so that final representations of farther sub-sequences are more global and coarse-grained. Pooling operations also reduce the dimension of sequence length, which makes parameter increasing rate controllable with the prolonging of input sequence. As a result, focal input sequence decomposition can also efficiently extract hierarchical features maps of input sequence when combined with other forecasting formulas with ICOM. Concrete architecture of feature extraction layer in Figure 7 is shown in Figure 8. We only use the canonical attention mechanism and maxpooling operation to perform experiments

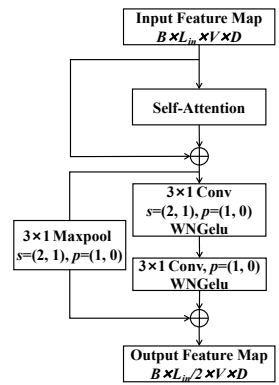

**Figure 8:** Architecture of feature extraction layer in Figure 7. Self-Attention is chosen as ICOM in this application and $s$ refers to stride.

---

[1] Additional pooling operations contain convolutional layers with stride=(2, 1) and maxpooling.

**Table 3**
Details of six datasets

| Dataset | Size | Dimension | Frequency |
|---------|------|-----------|-----------|
| ETTh$_1$ | 17420 | 7 | 1h |
| ETTm$_2$ | 69680 | 7 | 15min |
| ECL | 26304 | 321 | 1h |
| Traffic | 17544 | 862 | 1h |
| Exchange | 7588 | 8 | 1day |
| Weather | 52696 | 21 | 10min |

in later sections in order to emphasize the strength of focal input sequence decomposition.

## 6. Experiment

### 6.1. Datasets and Baselines

Extensive experiments are performed under six real-world datasets {ETTh$_1$, ETTm$_2$, ECL, Traffic, Exchange, Weather}[2]. Though ILI is also a commonly used real-world dataset, we abandon it in this paper according to two reasons below: (1) Compared with other six datasets, ILI dataset contains much fewer data points and numerical value of its timespan is only about 1000, which is dozens times smaller than those of other datasets. However, LSTI problem analyzed in this paper mainly tests forecasting capabilities of models when handling long input sequences. Obviously, ILI is too small to meet the requirement of long input sequences. (2) Compared with most of other models, FDNet owns weaker inductive bias, i.e., weaker constrains to model. Therefore, more instances are needed to help FDNet converge during training phase [56, 57]. Similarly, ILI is too small to meet the requirement of abundant instances. We will analyze more about these in the following Appendix C. We briefly introduce these six selected datasets and their usages in this paper below and their numeral details are shown in Table 3:

***ETT*** (Electricity Transformer Temperature) dataset, which consists of two versions subsets: 1-hour-level datasets {ETTh$_1$, ETTh$_2$} and 15-min-level datasets {ETTm$_1$, ETTm$_2$}, is composed of 2-years data of two separated electric stations in China. 'OT' (oil temperature) is the target value. For averting unnecessary experiments, we perform experiments on ETTh$_1$, a 1-hour-level subset, and ETTm$_2$, one 15-min-level subset, among these four subsets. The train/val/test is 12/4/4 months.

***ECL*** (Electricity Consuming Load) dataset contains the electricity consumption (Kwh) of 321 clients lasting for almost 2 years. It is converted into 2 years by informer [20]. 'MT_321' is set as the target value following the settings of FEDformer [22], ETSformer [8], etc. The train/val/test is 70%/10%/20%.

***Traffic*** dataset consists of road occupation rates in San Francisco Bay area freeways lasting for two years. It is collected hourly. The target is 'Sensor_861' and the train/val/test is 70%/10%/20%.

***Exchange*** consists of daily exchange rates in eight countries from 1990 to 2016. The target is set to 'Singapore' and the train/val/test is 70%/10%/20%.

***Weather*** dataset is a 10-min-level dataset which describes 21 meteorological indicators in Germany during 2020. The target variate name is set to 'OT' following FEDformer,ETSformer [8, 22]. The train/val/test is 70%/10%/20%.

Thirteen state-of-the-art time series forecasting models {FEDformer [22], ETSformer [8], Triformer [17], Crossformer [58], Non-stationary Transformer [59], PatchTST [60], SCINet [12], DLinear [47], N-HiTS [16], FiLM [61], Linear [62], AutoAI-TS [63], CoST [19]} are chosen as baselines. Specially, Crossformer is experimented only in multivariate forecasting conditions owing to its model traits.

### 6.2. Experiment Details

Hyper-parameter/Setting details of FDNet or other baselines within comparison experiments are shown in Table 4. They are fixed throughout entire experiments except specific instruction. The prediction length group is {96, 192, 336, 720}, which follows [8, 22]. MSE ($\sum_{i=1}^{n}(x_i-\hat{x}_i)^2/n$) and MAE ($\sum_{i=1}^{n}|x_i-\hat{x}_i|/n$) are chosen as metrics. All experiments are repeated 10 times and means of metrics are used. Results of seven baselines are directly borrowed from their papers if exist. We do their rest experiments according to their default settings. The best results are highlighted in bold and italic while the second best results are highlighted in underline and italic in all experiments except specific instruction. '—' denotes that models fail for out-of-memory (24GB) even when batch size = 1.

---

[2] Datasets were acquired at: https://drive.google.com/drive/folders/1ZOYpTUa82_jCcxIdTmyr0LXQfvaM9vIy?usp=sharing

**Table 4**
Details of hyper-parameters/settings

| Hyper-parameters/Settings | Values/Mechanisms |
|---|---|
| Input length | 672, 96 (only for Exchange) |
| $f$ | 5 |
| The number of decomposed feature extraction layers | 5 |
| Embedding dimension | 8 |
| The kernel size of Conv layers | $\{1 \times 1 \text{ (odd)}, 3 \times 1 \text{ (even)}\}$ |
| Standardization for data preprocessing | Z-score |
| Loss function | MSE |
| Optimizer | Adam |
| Activation | Gelu |
| Dropout | 0.1 |
| Learning rate | 1e-4 |
| Learning rate decreasing rate | Half per epoch |
| Batch size | 16 |
| Random seed | 4321 (if used) |
| Platform | Python 3.8.8 Pytorch 1.11.0 |
| Device | A single NVIDIA GeForce RTX 3090 24GB GPU |

## 6.3. Experiment Arrangement

We conduct diverse experiments to empirically demonstrate the superior forecasting performances of FDNet, and functions of its components and concepts:

1. Section 6.4 shows full results of multivariate/univariate forecasting results of all baselines and FDNet.
2. Section 6.5 takes CoST as an example to illustrate the problem of ICOM and universal feature extraction strategies. Results show that it is rational for FDNet to employ only the local/decomposed feature extraction strategy.
3. Section 6.6 performs ablation study to evaluate functions of FDNet components.
4. Section 6.7 performs experiments on Focal input sequence decomposition method to show that it is adaptive, available and beneficial for models with ICOM when handling rare forecasting conditions where universal feature maps and ICOM are feasible.
5. Section 6.8 compare LSTI handling capabilities of FDNet and other relatively better baselines in different settings.
6. Section 6.9 provides parameter sensitivity analysis of several hyper-parameters of FDNet.
7. Section 6.10 presents more experiment results on Exchange dataset, which is not very suitable for FDNet as settings of FDNet on Exchange dataset are different from other datasets.
8. Section 6.11 and Section 6.12 present showcases of KS Test experiment, which is mentioned in Section 4.4, and final representations of FDNet after T-SNE.

## 6.4. Main Results

We perform multivariate/univariate forecasting experiments to compare the forecasting capability of FDNet with those of mentioned baselines under six datasets. Results of FEDformer are average results of its two versions {FEDformer-f, FEDformer-w} and multivariate forecasting results of DLinear are average results of its two versions {DLinear-S, DLinear-I}. Input lengths of FDNet are all set to 672 and input sequences are all divided into 5 parts ($f = 5$) by focal decomposition method for all datasets except Exchange. As for Exchange dataset, the input length is set to 96 without focal decomposition method. Shorter input length is chosen for Exchange due to two reasons: (1) Exchange is a relatively small dataset compared with other five datasets. Though it is much bigger than ILI, 672 is nearly 1/10 of its entire size. (2) Input length of 96 is widely used by other selected baselines for Exchange. We will analyze more about forecasting performances of FDNet and other baselines with longer prediction length under Exchange in Section 6.10. Multivariate/Univariate forecasting results are respectively shown in Table 5/6.

**Table 5**
Results of multivariate forecasting

| Methods | Metrics | ETTh₁ | | | | ETTm₂ | | | | ECL | | | |
|---|---|---|---|---|---|---|---|---|---|---|---|---|---|
| | | 96 | 192 | 336 | 720 | 96 | 192 | 336 | 720 | 96 | 192 | 336 | 720 |
| FDNet | MSE | **0.365** | **0.400** | **0.427** | **0.457** | **0.168** | **0.237** | 0.310 | _0.417_ | **0.142** | **0.155** | **0.170** | **0.204** |
| | MAE | **0.397** | **0.419** | **0.438** | **0.482** | _0.260_ | _0.316_ | 0.369 | 0.437 | **0.242** | **0.254** | **0.271** | **0.301** |
| FEDformer | MSE | 0.419 | 0.461 | 0.530 | 0.686 | 0.204 | 0.293 | 0.342 | 0.427 | 0.188 | 0.198 | 0.213 | 0.239 |
| | MAE | 0.459 | 0.483 | 0.523 | 0.606 | 0.288 | 0.346 | 0.377 | 0.424 | 0.303 | 0.312 | 0.321 | 0.349 |
| ETSformer | MSE | 0.511 | 0.561 | 0.599 | 0.588 | 0.189 | 0.253 | 0.314 | 0.414 | 0.187 | 0.199 | 0.212 | 0.233 |
| | MAE | 0.487 | 0.513 | 0.529 | 0.541 | 0.280 | 0.319 | _0.357_ | 0.413 | 0.304 | 0.315 | 0.329 | 0.345 |
| Triformer | MSE | 0.419 | 0.484 | 0.513 | 0.711 | 0.240 | 0.387 | 0.545 | 1.928 | — | — | — | — |
| | MAE | 0.446 | 0.486 | 0.489 | 0.638 | 0.326 | 0.449 | 0.532 | 0.924 | — | — | — | — |
| Crossformer | MSE | 0.419 | 0.560 | 0.739 | 0.919 | 0.366 | 0.589 | 1.012 | 3.314 | 0.214 | 0.285 | 0.359 | 0.494 |
| | MAE | 0.437 | 0.525 | 0.635 | 0.747 | 0.432 | 0.568 | 0.725 | 1.294 | 0.301 | 0.350 | 0.386 | 0.453 |
| Non-stationary Transformer | MSE | 0.572 | 0.642 | 0.730 | 0.732 | 0.265 | 0.396 | 0.746 | 1.210 | 0.171 | 0.186 | 0.198 | _0.220_ |
| | MAE | 0.516 | 0.536 | 0.593 | 0.608 | 0.325 | 0.386 | 0.559 | 0.715 | 0.276 | 0.290 | 0.300 | _0.315_ |
| PatchTST | MSE | 0.393 | 0.438 | 0.482 | _0.492_ | 0.177 | _0.243_ | _0.305_ | 0.401 | 0.196 | 0.199 | 0.214 | 0.256 |
| | MAE | 0.427 | 0.459 | 0.494 | 0.519 | 0.274 | 0.324 | 0.363 | 0.420 | 0.279 | 0.283 | 0.299 | 0.331 |
| SCINet | MSE | 0.531 | 0.535 | 0.584 | 0.685 | 0.312 | 0.573 | 1.870 | 3.462 | 0.210 | 0.234 | 0.227 | 0.269 |
| | MAE | 0.503 | 0.513 | 0.560 | 0.634 | 0.415 | 0.591 | 1.078 | 1.753 | 0.333 | 0.345 | 0.340 | 0.373 |
| DLinear | MSE | _0.381_ | _0.429_ | _0.470_ | 0.513 | 0.193 | 0.265 | 0.331 | 0.418 | 0.189 | 0.189 | 0.202 | 0.238 |
| | MAE | _0.410_ | _0.440_ | _0.462_ | _0.512_ | 0.280 | 0.327 | 0.367 | 0.419 | 0.273 | 0.277 | 0.293 | 0.326 |
| N-HiTS | MSE | 0.422 | 0.476 | 0.527 | 0.581 | _0.176_ | 0.245 | **0.295** | **0.401** | _0.147_ | _0.167_ | _0.186_ | 0.243 |
| | MAE | 0.428 | 0.459 | 0.493 | 0.551 | **0.255** | **0.305** | **0.346** | _0.416_ | _0.249_ | _0.269_ | _0.290_ | 0.340 |
| FiLM | MSE | 0.700 | 0.718 | 0.721 | 0.722 | 0.230 | 0.285 | 0.339 | 0.433 | 0.845 | 0.849 | 0.861 | 0.891 |
| | MAE | 0.555 | 0.570 | 0.579 | 0.604 | 0.307 | 0.338 | 0.370 | 0.420 | 0.761 | 0.761 | 0.764 | 0.774 |
| Linear | MSE | 0.482 | 0.530 | 0.567 | 0.582 | 0.267 | 0.446 | 0.647 | 1.016 | 0.232 | 0.231 | 0.243 | 0.278 |
| | MAE | 0.471 | 0.497 | 0.516 | 0.548 | 0.346 | 0.443 | 0.535 | 0.675 | 0.330 | 0.331 | 0.343 | 0.371 |
| AutoAI-TS | MSE | 0.754 | 0.780 | 0.789 | 0.783 | 0.246 | 0.311 | 0.378 | 0.479 | 0.990 | 0.992 | 0.996 | 1.004 |
| | MAE | 0.594 | 0.606 | 0.618 | 0.638 | 0.322 | 0.358 | 0.396 | 0.447 | 0.832 | 0.832 | 0.833 | 0.834 |
| CoST | MSE | 0.499 | 0.652 | 0.804 | 0.973 | 0.289 | 0.509 | 0.800 | 1.657 | 0.163 | 0.172 | 0.196 | 0.232 |
| | MAE | 0.498 | 0.583 | 0.672 | 0.772 | 0.399 | 0.536 | 0.686 | 1.000 | 0.267 | 0.275 | 0.296 | 0.327 |

| Methods | Metrics | Traffic | | | | Exchange | | | | Weather | | | |
|---|---|---|---|---|---|---|---|---|---|---|---|---|---|
| | | 96 | 192 | 336 | 720 | 96 | 192 | 336 | 720 | 96 | 192 | 336 | 720 |
| FDNet | MSE | **0.402** | **0.412** | **0.424** | **0.466** | _0.085_ | 0.175 | 0.300 | _0.777_ | 0.159 | **0.200** | **0.247** | **0.309** |
| | MAE | **0.276** | **0.280** | **0.286** | **0.306** | 0.211 | 0.313 | 0.412 | _0.676_ | _0.211_ | **0.248** | **0.286** | **0.333** |
| FEDformer | MSE | 0.575 | 0.583 | 0.596 | 0.611 | 0.144 | 0.264 | 0.443 | 1.143 | 0.222 | 0.286 | 0.360 | 0.414 |
| | MAE | 0.358 | 0.360 | 0.353 | 0.375 | 0.277 | 0.375 | 0.482 | 0.821 | 0.300 | 0.350 | 0.398 | 0.431 |
| ETSformer | MSE | 0.607 | 0.621 | 0.622 | 0.632 | 0.085 | 0.182 | 0.348 | 1.025 | 0.197 | 0.237 | 0.298 | 0.352 |
| | MAE | 0.392 | 0.399 | 0.396 | 0.396 | _0.204_ | 0.303 | 0.428 | 0.774 | 0.281 | 0.312 | 0.353 | 0.388 |
| Triformer | MSE | — | — | — | — | 0.330 | 0.750 | 1.776 | 1.844 | 0.174 | 0.219 | 0.272 | 0.357 |
| | MAE | — | — | — | — | 0.406 | 0.611 | 0.966 | 0.986 | 0.242 | 0.290 | 0.323 | 0.378 |
| Crossformer | MSE | 0.535 | 0.546 | 0.553 | 0.571 | 0.328 | 0.691 | 0.931 | 1.526 | **0.155** | _0.207_ | _0.265_ | 0.409 |
| | MAE | 0.285 | 0.286 | 0.303 | _0.321_ | 0.431 | 0.637 | 0.765 | 0.962 | 0.228 | 0.277 | 0.322 | 0.404 |
| Non-stationary Transformer | MSE | 0.610 | 0.628 | 0.636 | 0.654 | 0.126 | 0.242 | 0.489 | 1.254 | 0.180 | 0.273 | 0.360 | 0.388 |
| | MAE | 0.337 | 0.347 | 0.348 | 0.357 | 0.249 | 0.356 | 0.506 | 0.821 | 0.228 | 0.304 | 0.364 | 0.380 |
| PatchTST | MSE | 0.589 | 0.579 | 0.589 | 0.631 | **0.080** | _0.170_ | 0.317 | 0.827 | 0.188 | 0.231 | 0.285 | 0.359 |
| | MAE | 0.379 | 0.368 | 0.371 | 0.391 | **0.197** | **0.293** | _0.405_ | 0.683 | 0.228 | 0.263 | 0.301 | _0.349_ |
| SCINet | MSE | 0.581 | 0.595 | — | — | 0.221 | 0.323 | 0.661 | 2.691 | 0.179 | 0.230 | 0.280 | 0.358 |
| | MAE | 0.423 | 0.429 | — | — | 0.365 | 0.442 | 0.564 | 1.320 | 0.255 | 0.299 | 0.331 | 0.353 |
| DLinear | MSE | 0.649 | 0.600 | 0.606 | 0.646 | _0.081_ | 0.158 | _0.255_ | **0.592** | 0.180 | 0.223 | 0.273 | _0.342_ |
| | MAE | 0.400 | 0.373 | 0.375 | 0.396 | 0.207 | _0.295_ | **0.385** | **0.604** | 0.246 | 0.289 | 0.331 | 0.381 |
| N-HiTS | MSE | _0.402_ | _0.420_ | _0.448_ | _0.539_ | 0.092 | 0.208 | 0.341 | 0.888 | _0.158_ | 0.211 | 0.274 | 0.351 |
| | MAE | _0.282_ | _0.297_ | _0.313_ | 0.353 | 0.211 | 0.322 | 0.422 | 0.723 | **0.195** | _0.247_ | _0.300_ | 0.353 |
| FiLM | MSE | 1.409 | 1.412 | 1.428 | 1.451 | 0.141 | 0.216 | 0.351 | 0.938 | 0.212 | 0.259 | 0.307 | 0.373 |
| | MAE | 0.799 | 0.802 | 0.806 | 0.809 | 0.272 | 0.342 | 0.434 | 0.736 | 0.263 | 0.297 | 0.326 | 0.367 |
| Linear | MSE | 0.848 | 0.805 | 0.812 | 0.852 | 0.118 | 0.189 | _0.282_ | 0.798 | 0.176 | 0.221 | 0.273 | 0.346 |
| | MAE | 0.513 | 0.495 | 0.498 | 0.514 | 0.261 | 0.335 | 0.413 | 0.701 | 0.253 | 0.296 | 0.335 | 0.387 |
| AutoAI-TS | MSE | 0.456 | 0.463 | 0.557 | 0.763 | 0.110 | 0.205 | 0.352 | 0.900 | 0.302 | 0.347 | 0.404 | 0.480 |
| | MAE | 0.487 | 0.499 | 0.555 | 0.652 | 0.236 | 0.325 | 0.432 | 0.719 | 0.289 | 0.317 | 0.353 | 0.399 |
| CoST | MSE | 0.453 | 0.459 | — | — | 0.259 | 0.467 | 0.853 | 1.124 | 0.355 | 0.501 | 0.654 | 0.884 |
| | MAE | 0.330 | 0.327 | — | — | 0.383 | 0.514 | 0.688 | 0.879 | 0.410 | 0.507 | 0.598 | 0.717 |

It could be observed from Table 5/6 that FDNet surpasses other baselines in most of situations. When compared with FEDformer/ ETSformer/ Triformer/ Crossformer/ Non-stationary Transformer/ PatchTST/ SCINet/ DLinear/ N-HiTS/FiLM/Linear/AutoAI-TS/CoST, FDNet yields 23.9%/18.2%/37.8%/53.7%/37.8%/14.3%/38.7%/9.8%/8.3%/

**Table 6**
Results of univariate forecasting

| Methods | Metrics | ETTh₁ |  |  |  | ETTm₂ |  |  |  | ECL |  |  |  |
|---|---|---|---|---|---|---|---|---|---|---|---|---|---|
| | | 96 | 192 | 336 | 720 | 96 | 192 | 336 | 720 | 96 | 192 | 336 | 720 |
| FDNet | MSE | 0.067 | 0.084 | 0.099 | 0.167 | 0.069 | ***0.099*** | ***0.129*** | ***0.173*** | ***0.203*** | *0.236* | *0.267* | *0.305* |
| | MAE | 0.200 | 0.226 | 0.248 | 0.331 | 0.196 | 0.241 | *0.277* | ***0.325*** | ***0.313*** | *0.337* | *0.362* | *0.409* |
| FEDformer | MSE | 0.115 | 0.137 | 0.142 | 0.144 | ***0.068*** | 0.106 | 0.139 | 0.199 | 0.258 | 0.299 | 0.354 | 0.435 |
| | MAE | 0.266 | 0.292 | 0.295 | 0.302 | 0.198 | 0.249 | 0.290 | 0.347 | 0.374 | 0.398 | 0.438 | 0.493 |
| ETSformer | MSE | *0.060* | *0.081* | *0.098* | 0.119 | 0.080 | 0.110 | 0.136 | *0.185* | 0.726 | 0.667 | 0.770 | 0.766 |
| | MAE | *0.190* | *0.221* | *0.248* | 0.282 | 0.213 | 0.252 | 0.283 | *0.333* | 0.656 | 0.625 | 0.677 | 0.674 |
| Triformer | MSE | 0.153 | 0.177 | 0.169 | 0.271 | 0.083 | 0.124 | 0.157 | 0.269 | 0.358 | 0.360 | 0.399 | 0.446 |
| | MAE | 0.324 | 0.347 | 0.336 | 0.453 | 0.221 | 0.271 | 0.310 | 0.408 | 0.424 | 0.433 | 0.456 | 0.499 |
| Non-stationary | MSE | 0.069 | 0.098 | 0.103 | 0.119 | 0.075 | 0.102 | 0.155 | 0.219 | 0.311 | 0.350 | 0.367 | 0.380 |
| Transformer | MAE | 0.198 | 0.241 | 0.250 | 0.275 | 0.203 | 0.241 | 0.300 | 0.375 | 0.415 | 0.434 | 0.449 | 0.461 |
| PatchTST | MSE | ***0.056*** | ***0.075*** | ***0.087*** | ***0.093*** | *0.068* | *0.101* | 0.132 | 0.183 | 0.319 | 0.332 | 0.397 | 0.450 |
| | MAE | ***0.179*** | ***0.208*** | ***0.230*** | ***0.240*** | *0.185* | *0.236* | *0.276* | 0.332 | 0.406 | 0.409 | 0.452 | 0.493 |
| SCINet | MSE | 0.119 | 0.129 | 0.160 | 0.243 | 0.076 | 0.102 | *0.129* | *0.176* | 0.312 | 0.314 | 0.332 | 0.364 |
| | MAE | 0.269 | 0.280 | 0.322 | 0.414 | 0.210 | 0.248 | 0.280 | *0.328* | 0.411 | 0.416 | 0.427 | 0.451 |
| DLinear | MSE | 0.067 | 0.088 | 0.109 | 0.191 | 0.072 | 0.104 | *0.136* | 0.189 | 0.391 | 0.369 | 0.396 | 0.434 |
| | MAE | 0.193 | 0.222 | 0.257 | 0.361 | *0.195* | *0.239* | *0.279* | 0.334 | 0.453 | 0.438 | 0.456 | 0.490 |
| N-HiTS | MSE | 0.125 | 0.146 | 0.176 | 0.258 | 0.072 | 0.109 | 0.150 | 0.192 | 0.342 | 0.356 | 0.400 | 0.458 |
| | MAE | 0.281 | 0.305 | 0.342 | 0.432 | 0.196 | 0.247 | 0.299 | 0.344 | 0.416 | 0.423 | 0.451 | 0.496 |
| FiLM | MSE | 0.066 | 0.083 | 0.097 | *0.102* | 0.152 | 0.175 | 0.197 | 0.238 | 0.967 | 0.958 | 0.992 | 1.038 |
| | MAE | 0.199 | 0.225 | 0.248 | *0.252* | 0.304 | 0.325 | 0.347 | 0.387 | 0.795 | 0.788 | 0.799 | 0.818 |
| Linear | MSE | 0.077 | 0.098 | 0.123 | 0.227 | 0.079 | 0.111 | 0.141 | 0.194 | 0.422 | 0.397 | 0.423 | 0.460 |
| | MAE | 0.208 | 0.237 | 0.275 | 0.396 | 0.210 | 0.250 | 0.287 | 0.340 | 0.477 | 0.460 | 0.476 | 0.508 |
| AutoAI-TS | MSE | 0.462 | 0.497 | 0.535 | 0.601 | 0.203 | 0.209 | 0.260 | 0.314 | 0.459 | 0.466 | 0.530 | 0.685 |
| | MAE | 0.593 | 0.616 | 0.644 | 0.700 | 0.356 | 0.359 | 0.402 | 0.451 | 0.501 | 0.502 | 0.538 | 0.631 |
| CoST | MSE | 0.080 | 0.104 | 0.121 | 0.302 | 0.076 | 0.112 | 0.145 | 0.216 | *0.208* | ***0.233*** | ***0.259*** | ***0.267*** |
| | MAE | 0.214 | 0.247 | 0.268 | 0.485 | 0.203 | 0.254 | 0.295 | 0.348 | *0.329* | *0.348* | *0.368* | ***0.379*** |

| Methods | Metrics | Traffic |  |  |  | Exchange |  |  |  | Weather |  |  |  |
|---|---|---|---|---|---|---|---|---|---|---|---|---|---|
| | | 96 | 192 | 336 | 720 | 96 | 192 | 336 | 720 | 96 | 192 | 336 | 720 |
| FDNet | MSE | ***0.134*** | ***0.135*** | ***0.135*** | ***0.162*** | *0.100* | ***0.186*** | 0.369 | ***0.652*** | 0.002 | 0.002 | ***0.002*** | ***0.003*** |
| | MAE | ***0.222*** | ***0.224*** | ***0.227*** | ***0.259*** | *0.245* | ***0.344*** | 0.479 | ***0.631*** | 0.030 | 0.034 | ***0.033*** | ***0.037*** |
| FEDformer | MSE | 0.189 | 0.189 | 0.199 | 0.216 | 0.143 | 0.282 | 0.469 | 1.232 | 0.005 | 0.006 | 0.006 | 0.010 |
| | MAE | 0.288 | 0.289 | 0.295 | 0.315 | 0.294 | 0.420 | 0.533 | 0.856 | 0.054 | 0.061 | 0.061 | 0.075 |
| ETSformer | MSE | 0.243 | 0.241 | 0.240 | 0.252 | 0.100 | 0.226 | 0.434 | 0.990 | 0.008 | 0.006 | 0.006 | 0.005 |
| | MAE | 0.355 | 0.352 | 0.353 | 0.362 | 0.252 | 0.353 | *0.500* | 0.821 | 0.078 | 0.065 | 0.065 | 0.062 |
| Triformer | MSE | 0.285 | 0.284 | 0.308 | 0.405 | 0.393 | 1.255 | 2.025 | 2.074 | 0.004 | 0.006 | 0.006 | 0.007 |
| | MAE | 0.367 | 0.363 | 0.383 | 0.457 | 0.505 | 0.927 | 1.194 | 1.105 | 0.050 | 0.062 | 0.063 | 0.070 |
| Non-stationary | MSE | 0.190 | 0.192 | 0.198 | 0.251 | 0.145 | 0.312 | *0.349* | 0.683 | 0.002 | ***0.002*** | 0.003 | *0.003* |
| Transformer | MAE | 0.273 | 0.287 | 0.294 | 0.352 | 0.301 | 0.418 | *0.462* | *0.650* | 0.028 | ***0.028*** | 0.036 | *0.038* |
| PatchTST | MSE | 0.280 | 0.227 | 0.221 | 0.244 | ***0.095*** | 0.203 | 0.399 | 1.032 | ***0.002*** | *0.002* | *0.003* | 0.004 |
| | MAE | 0.357 | 0.319 | 0.318 | 0.338 | ***0.230*** | *0.348* | 0.485 | 0.775 | ***0.025*** | *0.028* | *0.035* | 0.039 |
| SCINet | MSE | 0.217 | 0.299 | 0.259 | 0.278 | 0.209 | 0.347 | 0.575 | 1.378 | 0.008 | 0.008 | 0.008 | 0.008 |
| | MAE | 0.330 | 0.397 | 0.365 | 0.379 | 0.366 | 0.475 | 0.604 | 0.939 | 0.068 | 0.069 | 0.070 | 0.071 |
| DLinear | MSE | 0.361 | 0.309 | 0.305 | 0.351 | 0.118 | *0.222* | *0.400* | 0.837 | 0.006 | 0.006 | 0.006 | 0.007 |
| | MAE | 0.442 | 0.395 | 0.392 | 0.425 | 0.277 | 0.382 | 0.506 | 0.722 | 0.062 | 0.066 | 0.067 | 0.069 |
| N-HiTS | MSE | 0.284 | 0.264 | 0.269 | 0.298 | 0.241 | 0.870 | 1.809 | 2.144 | 0.003 | 0.004 | 0.003 | 0.004 |
| | MAE | 0.369 | 0.354 | 0.361 | 0.384 | 0.372 | 0.700 | 1.061 | 1.152 | 0.041 | 0.050 | 0.046 | 0.048 |
| FiLM | MSE | 1.861 | 1.845 | 1.836 | 1.828 | 0.152 | 0.256 | 0.463 | 1.002 | *0.002* | 0.003 | 0.004 | 0.004 |
| | MAE | 1.173 | 1.169 | 1.167 | 1.163 | 0.308 | 0.406 | 0.531 | 0.774 | *0.027* | 0.029 | 0.038 | 0.039 |
| Linear | MSE | 0.475 | 0.430 | 0.430 | 0.475 | 0.135 | 0.241 | 0.435 | 0.860 | 0.006 | 0.006 | 0.007 | 0.007 |
| | MAE | 0.519 | 0.484 | 0.485 | 0.513 | 0.300 | 0.400 | 0.526 | 0.735 | 0.064 | 0.066 | 0.068 | 0.070 |
| AutoAI-TS | MSE | 0.393 | 0.390 | 0.464 | 0.652 | 0.118 | *0.190* | ***0.308*** | *0.669* | 0.002 | 0.004 | 0.004 | 0.002 |
| | MAE | 0.456 | 0.454 | 0.503 | 0.601 | 0.284 | 0.364 | ***0.459*** | 0.660 | 0.028 | 0.050 | 0.031 | 0.035 |
| CoST | MSE | *0.156* | *0.158* | *0.163* | *0.182* | 0.107 | 0.225 | 0.431 | 0.778 | 0.010 | 0.009 | 0.009 | 0.009 |
| | MAE | *0.243* | *0.245* | *0.252* | *0.268* | 0.263 | 0.381 | 0.512 | 0.682 | 0.077 | 0.073 | 0.073 | 0.075 |

50.1%/47.2%/46.5%/42.5% relative MSE reduction during multivariate forecasting and when compared with FEDformer/ETSformer/Triformer/Non-stationary Transformer/PatchTST/SCINet/DLinear/N-HiTS/FiLM/Linear/AutoAI-TS/CoST, FDNet yields 30.2%/30.6%/52.2%/19.2%/24.5%/37.5%/22.9%/31.1%/73.6%/39.6%/55.1%/40.4% relative MSE reduction during univariate forecasting in general, which shows the superior forecasting capability of FDNet.

Specially, performing univariate forecasting only on a specific variable of an arbitrary multivariate dataset cannot persuasively illustrate the univariate forecasting capability of a certain model. However, it is a commonly agreed-upon setting, albeit not very considerate, for recent researches to present univariate forecasting results by only using the last variable of every dataset. To more persuasively present the univariate forecasting capability of FDNet, we supplement univariate forecasting results with M4 [64] in Appendix C.2.

**Table 7**
Results of CoST with different kernel sizes during Weather univariate forecasting

| Pred | 96 | | | 192 | | | 336 | | | 720 | | |
|---|---|---|---|---|---|---|---|---|---|---|---|---|
| Kernel size | MSE | MAE | Rank | MSE | MAE | Rank | MSE | MAE | Rank | MSE | MAE | Rank |
| {1, 2, 4, 8, 16, 32, 64, 128} | 0.010 | 0.077 | 13 | 0.009 | 0.073 | 13 | 0.009 | 0.073 | 13 | 0.009 | 0.075 | 12 |
| {1, 2, 4, 8, 16, 32, 64} | 0.006 | 0.062 | 11 | 0.006 | 0.061 | 7 | 0.006 | 0.062 | 7 | 0.007 | 0.066 | 8 |
| {1, 2, 4, 8, 16, 32} | 0.006 | 0.060 | 11 | 0.006 | 0.059 | 7 | 0.006 | 0.060 | 7 | 0.006 | 0.064 | 8 |
| {1, 2, 4, 8, 16} | _0.005_ | _0.058_ | _8_ | **0.005** | **0.058** | **7** | **0.006** | **0.060** | **7** | **0.006** | **0.063** | **8** |
| {1, 2, 4, 8} | 0.006 | 0.060 | 11 | 0.006 | 0.060 | 7 | 0.006 | 0.061 | 7 | _0.006_ | _0.064_ | _8_ |
| {1, 2, 4} | 0.006 | 0.061 | 11 | 0.006 | 0.061 | 7 | 0.006 | 0.062 | 7 | 0.007 | 0.065 | 8 |
| {1, 2} | **0.005** | **0.057** | **8** | _0.006_ | _0.058_ | _7_ | _0.006_ | _0.061_ | _7_ | 0.007 | 0.065 | 8 |
| {32, 64, 128} | 0.011 | 0.081 | 13 | 0.010 | 0.079 | 13 | 0.010 | 0.080 | 13 | 0.010 | 0.079 | 12 |
| {64, 128} | 0.011 | 0.081 | 13 | 0.010 | 0.077 | 13 | 0.010 | 0.078 | 13 | 0.012 | 0.084 | 13 |
| {128} | 0.015 | 0.095 | 13 | 0.012 | 0.086 | 13 | 0.012 | 0.085 | 13 | 0.012 | 0.085 | 13 |

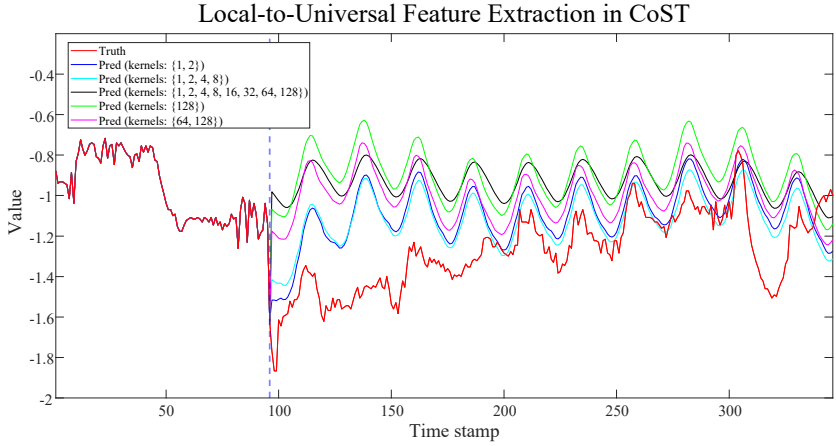

Figure 9: Visualization of typical results in Table 7. We visualize forecasting results of CoST with large kernels ({128}, {64, 128}), with small kernels ({1, 2}, {1, 2, 4, 8}) and its initial kernels ({1, 2, 4, 8, 16, 32, 64, 128}). The red line located in the left part of the blue dotted line is input sequence while lines in the right part of it are truth (red) and prediction results (other colors).

## 6.5. Universal/Local Feature Extraction Methods

Most of other baselines own ICOMs and intend to extract universal feature maps of input sequence, however their general performances are worse than that of FDNet extracting only local features in Table 5/6. It empirically demonstrates that decomposed forecasting formula is more suitable for real-world forecasting conditions. In other words, local feature extraction method seems more practical than the universal one in time series forecasting networks. Specially, CoST [19] extracts global-local feature maps of input sequence through a mixture of experts owning different convolution kernel sizes within {1, 2, 4, 8, 16, 32, 64, 128}. However, its performance is still worse than that of FDNet, indicating that universal feature maps might bring pernicious effects. To validate this statement, we *gradually remove its large kernels within its convolution kernel group* to redo univariate forecasting experiment under Weather dataset. Forecasting results with only large kernels are also presented. Results are shown in Table 7. It could be observed that the forecasting accuracy of CoST rises with smaller kernels and its forecasting rank rises synchronously among all fourteen methods after our transformation. To visualize results in Table 7, we provide a typical forecasting condition with non-stationary sequence in Figure 9. It can be observed that forecasting performances of CoST with larger kernels are more vulnerable to distribution shifts. Specially, CoST only using large kernels own the worst performances and most severe over-fitting problem. Some forecasting curves of CoST with large kernels even lose the continuity of time series sequences, resulting in catastrophic forecasting errors. As smaller kernels lead to more local receptive fields, this result once more demonstrates that extracting local fine-grained features is more practical and useful for real-world time series forecasting task.

**Table 8**
Results of ablation study on focal decomposition method and usage of convolutions

| Focal | Traffic (Univariate) | | | | | | | | Traffic (Multivariate) | | | | | | | |
|---|---|---|---|---|---|---|---|---|---|---|---|---|---|---|---|---|
| | 96 | | 192 | | 336 | | 720 | | 96 | | 192 | | 336 | | 720 | |
| | MSE | MAE | MSE | MAE | MSE | MAE | MSE | MAE | MSE | MAE | MSE | MAE | MSE | MAE | MSE | MAE |
| Focal | **0.133** | *0.222* | **0.135** | *0.224* | **0.135** | **0.227** | **0.162** | *0.259* | **0.402** | **0.276** | **0.412** | **0.280** | **0.424** | **0.286** | **0.466** | **0.306** |
| Pyramid | *0.134* | **0.215** | **0.135** | **0.218** | *0.136* | **0.225** | **0.162** | **0.256** | 0.418 | 0.279 | 0.430 | 0.284 | 0.441 | 0.288 | 0.476 | *0.306* |
| Patch | 0.138 | 0.230 | 0.138 | 0.228 | 0.142 | 0.235 | 0.165 | 0.261 | *0.406* | *0.279* | *0.415* | *0.282* | *0.428* | *0.288* | *0.470* | 0.308 |

| Conv (1-st dimension) | Weather (Univariate) | | | | | | | | Weather (Multivariate) | | | | | | | |
|---|---|---|---|---|---|---|---|---|---|---|---|---|---|---|---|---|
| | 96 | | 192 | | 336 | | 720 | | 96 | | 192 | | 336 | | 720 | |
| | MSE | MAE | MSE | MAE | MSE | MAE | MSE | MAE | MSE | MAE | MSE | MAE | MSE | MAE | MSE | MAE |
| Conv | **1.8e-3** | **0.030** | **2.2e-3** | **0.034** | **2.1e-3** | **0.033** | **2.6e-3** | **0.037** | **0.159** | **0.211** | **0.200** | **0.248** | **0.247** | **0.286** | **0.309** | **0.333** |
| Linear | *2.4e-3* | *0.036* | *2.5e-3* | *0.037* | *2.7e-3* | *0.039* | *3.0e-3* | *0.040* | *0.160* | *0.213* | *0.201* | *0.248* | *0.248* | *0.287* | *0.310* | *0.334* |

| Conv (2-nd dimension) | Weather (Multivariate) | | | | | | | |
|---|---|---|---|---|---|---|---|---|
| | 96 | | 96 | | 96 | | 96 | |
| | MSE | MSE | MSE | MSE | MSE | MSE | MSE | MSE |
| Ours | *0.159* | *0.211* | *0.200* | *0.248* | *0.247* | *0.286* | *0.309* | *0.333* |
| FDNet-A | 0.170 | 0.229 | 0.213 | 0.271 | 0.258 | 0.308 | 0.318 | 0.358 |
| FDNet-S | **0.157** | *0.215* | *0.204* | *0.256* | *0.250* | *0.300* | *0.314* | *0.352* |

**Table 9**
Results of FUNet under ETTh$_1$

| Methods | ETTh$_1$ (Univariate) | | | | | | | | ETTh$_1$ (Multivariate) | | | | | | | |
|---|---|---|---|---|---|---|---|---|---|---|---|---|---|---|---|---|
| | 96 | | 192 | | 336 | | 720 | | 96 | | 192 | | 336 | | 720 | |
| | MSE | MAE | MSE | MAE | MSE | MAE | MSE | MAE | MSE | MAE | MSE | MAE | MSE | MAE | MSE | MAE |
| FUNet | **0.059** | **0.190** | **0.071** | **0.202** | **0.088** | **0.230** | **0.112** | **0.263** | *0.402* | *0.416* | *0.458* | *0.448* | *0.501* | *0.473* | *0.582* | 0.550 |
| FDNet | *0.067* | *0.200* | *0.084* | *0.226* | *0.099* | *0.248* | 0.167 | 0.331 | **0.365** | **0.397** | **0.400** | **0.419** | **0.427** | **0.438** | **0.457** | **0.482** |

## 6.6. Ablation Study

We perform ablation study on focal decomposition method and $3 \times 1$ convolutional layer used in decomposed feature extraction layer to verify their corresponding functions.

As for focal decomposition method, three ablation variants are tested under Traffic dataset: (1) Focal: FDNet with focal decomposition method; (2) Pyramid: FDNet with pyramid decomposition method [20]; (3) Patch: FDNet with patch decomposition method [17]. Number of pyramids/patches are set as 3/4 following their default settings. Experiment results are shown in the upper part of Table 8. It could be observed that focal decomposition method behaves better in most of experiments compared with other two decomposition methods, showing that it is more competitive in forecasting tasks.

As for the usage of convolutions, their kernels own two dimensions, whose functions are different, so that the functions of two dimensions need to be tested repsectively. Two variants are tested under Weather dataset to evaluate the function of the kernel in the temporal dimension (1-st dimension): (1) Conv: Initial FDNet in Figure 4; (2) Linear: FDNet with decomposed feature extractor layers containing only $1 \times 1$ convolutional layers. The experiment results are shown in the middle part of Table 8. It could be observed that the forecasting performances of FDNet drop obviously without the usage of convolutions, especially when handling the univariate forecasting conditions. Though the function of the convolution kernel in the temporal dimension is diluted in multivariate forecasting conditions, where the intra-relationship and the inter-relationship both count, it cannot be denied that the convolution kernel in the temporal dimension is still beneficial. Another three variants are tested under the same dataset to evaluate the function of the kernel in the variable dimension (2-nd dimension): (1) Ours: Initial FDNet in Figure 4; (2) FDNet-A: FDNet with variable-agnostic forecasting formula; (3) FDNet-S: FDNet with variable-specific forecasting formula. Note that 2D convolutions are no longer used in the later two variants of FDNet. The variates are treated as the features of the temporal dimension in FDNet-A while different networks are used for different variates in FDNet-S. Therefore, the 2D convolution degrades into 1D convolution with the kernel size of 3 in FDNet-A and FDNet-S. The experiment results are shown in the bottom part of Table 8. The performances of FDNet surpass both of two other variants in nearly all occasions, illustrating that our proposed 2D convolution have an edge on handling inter-relationships than the prevalent forecasting formulas. These phenomena empirically demonstrate the function of the specific 2D convolutions used in decomposed feature extraction layers.

## 6.7. Focal with ICOM

We perform experiments on FUNet shown in Figure 7 to validate that focal input sequence decomposition method with ICOM can compensate the shortcoming of FDNet when global feature maps exist and are beneficial. Experiments are conducted under univariate forecasting of ETTh$_1$ where FDNet is not so competitive with some selected baselines, e.g., ETSformer and FEDformer. Results are shown in Table 9. It shows that under those rare occasions where universal feature maps are needed, focal input sequence decomposed method could transform FDNet to FUNet to outperform those SOTA forecasting methods with elaborately designed ICOMs even only using the most basic ICOMs including canonical attention and convolution. We additionally supply multivariate forecasting results of FUNet in Table 9. Results with underline and italic here mean that they are worse than results of FDNet but better than results of other baselines in Table 5. It could be inferred that though FUNet performs better than other baselines, it fails to challenge FDNet in general. This once more illustrates that local feature extraction methods are more practical and general for time series forecasting and the outstanding forecasting capability of FDNet architecture regardless of armed with ICOM or not.

## 6.8. LSTI Problem Handling Capability

To verify that FDNet is more accurate, efficient and robust in handling LSTI problems, we conduct two group of experiments in this section. Two groups of experiments respectively compare comprehensive forecasting capabilities of FDNet with 'Complicated. and 'Simple' baselines under certain conditions owing to their different network properties. *Whether a baseline is 'Complicated' or 'Simple' here is determined by whether this baseline processes feature maps of input sequences in latent space or not.* Among seven baselines, N-HiTS and DLinear are 'Simple' baselines in that they only employ linear projection layers and abandon embedding processes while others including FDNet are 'Complicated' baselines based on Transformer/CNN.

### 6.8.1. FDNet vs. 'Complicated' Baselines

FEDformer-w and ETSformer are chosen as baselines due to their generally outstanding performances within 'Complicated' baselines in Table 5/6. This group of experiment is conducted under the univariate and multivariate forecasting of ECL. Input sequence lengths are chosen within {96, 672, 1344} and prediction sequence is set to 96. Batch size of univariate forecasting is 16 and 1 for multivariate forecasting (FEDformer-w will be out of memory with batch size of 16 in more than half of conditions). Input sequences of FDNet are respectively divided into {4, 5, 6} parts by focal decomposition method. Each sub-experiment is done for 20 times. Means and standard deviations (Stds) of all forecasting MSEs /MAEs/SMAPEs are shown in Table 10, together with their GPU memory occupations (GPU), average training time per epoch (ATPE), total inference time (TIT) and total learnable parameters (TLP). It could be observed from Table 10 that FDNet performs worse than FEDformer-w when it comes to the shortest input sequence condition. However, it performs better than other two baselines in the rest of conditions proving that FDNet is better at handling LSTI problem and more accurate. Specially, forecasting errors measured by three evaluation metrics of FDNet with input sequence lengths 672 or 1344 are only slightly different while those of FEDformer-w grow apparently when input sequence length changes from 672 to 1344. Moreover, Stds of FDNet are much lower than those of other two baselines in any forecasting condition, demonstrating the robustness of FDNet. What's more, except the shortest input sequence condition where ATPE of FDNet is bigger than that of ETSformer, all GPU/ATPE/TIT/TLP of FDNet are smaller than those of other baselines, illustrating better computation efficiency of FDNet. Meanwhile, Table 10 reveals a deficiency of FDNet, i.e., the forecasting performance of FDNet is worse than those of most of other models with relatively short input sequence length. The requirement of long input sequence length for extracting feature maps makes FDNet weak in handling small dataset. However, this requirement does not mean that FDNet obtains better forecasting performance than other models but pays the price of more model parameters. As Table 10 shows, when prolonging input sequence length from 96 to 1344, GPU of FDNet increases slightly. Moreover, even GPU of FDNet with input sequence length of 1344 is much smaller than those of FEDformer/ ETSformer with input sequence length of 96, which manifests the high efficiency of FDNet.

**Table 10**
Results on LSTI problem handling capability with 'Complicated' baselines

| Input length | Methods | ECL (Univariate) | | | | | | | | | |
|---|---|---|---|---|---|---|---|---|---|---|---|
| | | Mean (MSE) | Std (MSE) | Mean (MAE) | Std (MAE) | Mean/% (SMAPE) | Std/% (SMAPE) | GPU/MB | ATPE/s | TIT/s | TLP/MB |
| 96 | FDNet | *0.391* | **1.0e-3** | *0.463* | **9.0e-4** | *83.1* | **0.09** | **1516** | *35.2* | **2.0** | **0.1** |
| | ETSformer | 0.726 | *1.7e-3* | 0.786 | 6.5e-3 | 137.5 | 2.38 | 2252 | **26.3** | 3.4 | *5.0* |
| | FEDformer-w | **0.268** | 1.8e-3 | **0.434** | **2.0e-3** | **75.1** | *0.23* | 6445 | 444.0 | 19.5 | 116.1 |
| 672 | FDNet | **0.204** | **1.3e-3** | **0.310** | **9.4e-4** | **60.2** | **0.08** | **1535** | **65.3** | **5.0** | **0.5** |
| | ETSformer | 0.891 | 2.0e-3 | 0.767 | *3.8e-3* | 138.4 | 2.39 | 3506 | 68.3 | 5.7 | *5.0* |
| | FEDformer-w | *0.294* | *1.9e-3* | *0.443* | 9.1e-3 | *77.5* | *1.68* | 7885 | 583.2 | 25.6 | 116.1 |
| 1344 | FDNet | **0.209** | **1.5e-3** | **0.322** | **1.2e-3** | **62.0** | **0.05** | **1551** | **84.5** | **6.6** | **1.0** |
| | ETSformer | 0.893 | *3.9e-3* | 0.769 | *2.5e-3* | 135.8 | *0.51* | 4954 | 92.7 | 9.0 | *5.0* |
| | FEDformer-w | *0.323* | 1.1e-2 | *0.448* | 8.2e-3 | *78.0* | 1.61 | 10059 | 697.5 | 30.6 | 116.1 |
| Input length | Methods | ECL (Multivariate) | | | | | | | | | |
| | | Mean (MSE) | Std (MSE) | Mean (MAE) | Std (MAE) | Mean/% (SMAPE) | Std/% (SMAPE) | GPU/MB | ATPE/s | TIT/s | TLP/MB |
| 96 | FDNet | *0.215* | **3.1e-4** | **0.303** | **3.9e-4** | *57.5* | *0.06* | **1943** | **867.3** | **40.6** | **0.1** |
| | ETSformer | 0.832 | 4.5e-3 | 0.757 | *5.1e-3* | 144.7 | 0.12 | *2570* | *917.8* | *45.8* | *6.3* |
| | FEDformer-w | **0.193** | **2.9e-3** | *0.308* | 1.0e-3 | **57.0** | **0.04** | 6700 | 11577.8 | 674.3 | 118.1 |
| 672 | FDNet | **0.138** | **3.7e-4** | **0.237** | **4.2e-4** | **47.3** | 0.07 | **2062** | **873.1** | **45.9** | **0.5** |
| | ETSformer | 0.820 | 4.2e-3 | 0.752 | *2.5e-3* | 140.1 | **0.06** | *2836* | *1092.7* | *51.4* | *6.3* |
| | FEDformer-w | *0.208* | *3.9e-3* | *0.322* | 2.8e-3 | *58.4* | **0.06** | 6845 | 13434.6 | 782.5 | 118.1 |
| 1344 | FDNet | **0.139** | **3.3e-4** | **0.239** | **7.2e-4** | **47.8** | 0.09 | **2132** | **883.7** | **51.1** | **1.0** |
| | ETSformer | 0.820 | 3.5e-3 | 0.752 | 1.8e-3 | 139.9 | 0.12 | *3006* | *1284.8* | *56.8* | *6.3* |
| | FEDformer-w | *0.214* | **2.0e-3** | *0.327* | *1.3e-3* | *58.7* | *0.12* | 7011 | 14317.8 | 833.9 | 118.1 |

### 6.8.2. FDNet vs. 'Simple' Baselines

Comprehensive forecasting performances of N-HiTS [16] and DLinear [47] are better than all other baselines except FDNet. Therefore, we perform more comparison experiments, which are conducted under the multivariate forecasting of ECL and Traffic, in this group. We do not perform univariate forecasting here in that these two simple baselines have extremely little learning parameters in virtue of abandoning embedding processes. Obviously, FDNet has no chance in having less GPU or time complexity than these two models under univariate forecasting though its forecasting accuracy is better. What's more, discussing input length selection is also meaningless owing to the same reasons. However, in this experiment, we prolong input sequence lengths of both two baselines from 96 (their own default settings) to 672 (setting of FDNet) for fair input sequence circumstances. Besides, we remove focal decomposition method of FDNet, i.e., $f = 1$, to test whether FDNet could achieve better comprehensive forecasting perform-ances under multivariate forecasting without the help of focal decomposition method. The number of decomposed feature extraction layer is also set to 1 in that these two baselines has few layers, e.g. DLinear even has only one linear projection layer for each variate. Results are shown in Table 11. Evaluation metrics are identical to those used in comparison experiments with 'Complicated' baselines.

**Table 11**
Results on LSTI problem handling capability with 'Simple' baselines

| Pred | Methods | ECL (Multivariate) | | | | | | | | | |
|---|---|---|---|---|---|---|---|---|---|---|---|
| | | Mean (MSE) | Std (MSE) | Mean (MAE) | Std (MAE) | Mean (SMAPE)/% | Std (SMAPE)/% | GPU/MB | ATPE/s | TIT/s | TLP/MB |
| 96 | FDNet | **0.138** | 4.7e-4 | **0.233** | 9.6e-5 | **46.7** | 0.13 | 3106 | **46.7** | **10.1** | **1.0** |
| | DLinear | *0.143* | **6.0e-5** | 0.246 | *6.1e-4* | 49.0 | **0.09** | *1288* | *303.1* | *28.4* | 39.6 |
| | N-HiTS | 0.151 | 5.0e-4 | 0.258 | 1.2e-3 | 51.0 | *0.12* | **1258** | 1152.4 | 163.7 | *2.0* |
| 192 | FDNet | **0.152** | *3.3e-4* | **0.246** | 6.6e-4 | **48.4** | *0.10* | 3124 | **47.9** | **10.7** | **2.0** |
| | DLinear | *0.159* | **1.7e-4** | *0.263* | **2.8e-4** | *51.4* | 0.12 | *2576* | *303.2* | *29.8* | 79.1 |
| | N-HiTS | 0.172 | 5.0e-4 | 0.269 | 9.4e-4 | 52.4 | **0.10** | **2516** | 1169.2 | 166.9 | *2.1* |
| 336 | FDNet | **0.167** | 4.6e-4 | **0.262** | 5.6e-4 | **50.6** | *0.12* | **3147** | **48.8** | **11.4** | *3.4* |
| | DLinear | *0.175* | **2.8e-4** | 0.283 | 7.7e-4 | 54.2 | **0.09** | 4508 | *310.4* | *33.2* | 158.2 |
| | N-HiTS | 0.188 | 6.0e-4 | *0.280* | *7.5e-4* | *53.8* | 0.14 | *4403* | 1263.6 | 173.3 | **2.2** |
| 720 | FDNet | **0.201** | 9.4e-4 | **0.296** | 6.0e-4 | **55.1** | *0.13* | **3159** | **52.6** | **13.3** | *7.4* |
| | DLinear | *0.214* | **2.6e-4** | 0.315 | **1.3e-4** | 58.3 | **0.10** | 9660 | *310.8* | *39.8* | 341.4 |
| | N-HiTS | 0.243 | 1.0e-3 | *0.301* | 1.8e-3 | *56.4* | 0.32 | *9435* | 1295.7 | 189.4 | **2.4** |

| Pred | Methods | Traffic (Multivariate) | | | | | | | | | |
|---|---|---|---|---|---|---|---|---|---|---|---|
| | | Mean (MSE) | Std (MSE) | Mean (MAE) | Std (MAE) | Mean (SMAPE)/% | Std (SMAPE)/% | GPU/MB | ATPE/s | TIT/s | TLP/MB |
| 96 | FDNet | **0.396** | 4.7e-4 | **0.274** | 8.1e-4 | 49.3 | 0.17 | 5699 | **87.7** | **13.6** | **1.0** |
| | DLinear | 0.457 | **2.1e-4** | 0.349 | 1.2e-3 | 62.6 | *0.22* | *4385* | *484.8* | *47.0* | 106.2 |
| | N-HiTS | *0.411* | *4.0e-4* | *0.302* | *1.8e-3* | *54.4* | 0.29 | **1538** | 2120.5 | 293.1 | *2.0* |
| 192 | FDNet | **0.407** | *2.2e-4* | **0.276** | 8.0e-4 | 49.4 | **0.11** | *5732* | **90.6** | **14.6** | **2.0** |
| | DLinear | 0.469 | **1.8e-4** | 0.354 | **5.3e-4** | 63.2 | *0.14* | 8770 | *487.3* | *48.8* | 216.8 |
| | N-HiTS | *0.424* | 4.0e-4 | *0.310* | 6.5e-3 | *55.8* | 1.24 | **3076** | 2124.3 | 310.3 | *2.1* |
| 336 | FDNet | **0.418** | *3.1e-4* | **0.284** | 1.2e-3 | 50.5 | 0.17 | *5753* | **91.5** | **16.8** | *3.4* |
| | DLinear | 0.483 | **2.0e-4** | 0.360 | 7.2e-4 | 64.1 | **0.09** | 15348 | *487.5* | *48.9* | 442.1 |
| | N-HiTS | *0.453* | 5.0e-4 | *0.309* | 2.0e-3 | *55.3* | 0.27 | **5383** | 2137.2 | 317.2 | **2.2** |
| 720 | FDNet | **0.457** | 6.4e-4 | **0.303** | 9.0e-4 | 53.4 | 0.13 | 5819 | **96.5** | **20.7** | *7.4* |
| | DLinear | — | — | — | — | — | — | — | — | — | — |
| | N-HiTS | *0.491* | 5.0e-4 | *0.316* | 4.2e-3 | *56.3* | 0.52 | 11535 | 2159.5 | 340.4 | **2.4** |

It could be observed from Table 11 that: (1) Average forecasting errors evaluated by three metrics of FDNet are lower than those of DLinear and N-HiTS when prolonging their input sequence length to 672. Moreover, forecasting accuracy of DLinear and N-HiTS decrease in nearly all conditions when changing their input sequence length from 96 to 672. It means that FDNet behaves better than these two baselines from the prospective of forecasting accuracy when tackling LSTI problems. (2) It is commonly known that simpler methods are less likely to be affected by over-fitting problem, i.e., their robustness shall be brilliant. However, all three methods share similar standard deviations under every forecasting condition, showing that FDNet can achieve similar robustness with those simple methods even it is a relatively more complicated method. (3) DLinear and N-HiTS are both variable-specific methods, but they are trained and inferred in different ways. DLinear perfers to minimize the training/inference time so that it trains/infers windows of different variates simultaneously. As we have analyzed in Section 5.1 before, this behavior will bring unacceptable space occupation. Therefore, GPU and TLP of DLinear is much bigger than those of others in Table 11 especially with Traffic which owns 862 variates. N-HiTS are trained and inferred in the completely opposite way. N-HiTS trains/infers windows of different variates independently. Training/Inference process of a certain variate will only begin after corresponding process of the former variate ends. Therefore, though it has much smaller space occupation, it owns much longer training/inference time than other two methods. Different from these variable-specific methods, FDNet can achieve a trade-off between time and space complexity due to its application of specific 2D convolution. It could be seen that FDNet owns the smallest ATPE/TIT and the intermediate GPU/TLP. In addition, FDNet is the only method that avoids parameter explosion with the increase of prediction length. Though DLinear and N-HiTS may have smaller GPU when the prediction length is short, e.g., 96, GPU of them increase linearly with the prediction length while that of FDNet increase slightly. Thus, they all own more GPU when the prediction horizon is longer, e.g., 720. We must emphasize again that FDNet is a much more complicated model than DLinear and N-HiTS. However, FDNet could still achieve such comprehensively better performances when considering both forecasting accuracy and efficiency, demonstrating the successful design of our model architecture for multivariate forecasting.

Consequently, FDNet can deal with LSTI problems better than both 'Complicated' and 'Simple' baselines in general, demonstrating its feasibility and robustness in real-world applications.

## 6.9. Parameter sensitivity

We conduct experiment here on evaluating comprehensive performances of FDNet with respect to hyper-parameter $f$ during multivariate forecasting under ECL/Traffic. We set $f$ to the range of 1 to 5 and metrics include GPU and MSE. The number of decomposed feature extraction layer for the latest decomposed sub-sequence is always 5. Note that when $f = 1$, this certain network is equivalent to FDNet without focal decomposition method. Results are shown in Table 12. We only present GPU of FDNet under prediction length of 720 in that GPU of FDNet changes slightly with prolonging input sequence length according to analysis in Section 6.8.

It could be observed from Table 12 that: (1) Forecasting errors change slightly with different $f$ for two datasets under all prediction lengths, illustrating the robustness of decomposed forecasting formula. (2) With bigger $f$, GPU, which reveals the space complexity of the model, apparently decreases. This shows that focal decomposition method efficiently reduce the space complexity of FDNet without bringing distinct performance drop. (3) $f$ is set to 5 in our main results in Table 5, however, $f = 5$ is neither the optimal solution for ECL nor Traffic. In fact, in Table 5 $f$ could be adjusted to 3/2 for better results respectively for two datasets. It means that though forecasting performances of FDNet has already shown better results than other baselines with fixed $f$, it can even behave better if we fine-tune hyper-parameter $f$ for different forecasting conditions. (4) FDNet with $f = 3/2$ has best performances under all prediction lengths for ECL/Traffic, which shows the consistency of forecasting performances of FDNet with focal decomposition method.

**Table 12**
Parameter sensitivity experiment on $f$

| | ECL (Multivariate) | | | | | Traffic (Multivariate) | | | | |
|---|---|---|---|---|---|---|---|---|---|---|
| $f$/Metrics | 96 | 192 | 336 | 720 | | 96 | 192 | 336 | 720 | |
| | MSE | MSE | MSE | MSE | GPU/MB | MSE | MSE | MSE | MSE | GPU/MB |
| 1 | 0.136 | 0.150 | 0.165 | 0.200 | 7035 | _0.392_ | _0.403_ | _0.415_ | _0.455_ | 16415 |
| 2 | _0.135_ | _0.149_ | _0.165_ | **0.198** | 7051 | **0.388** | **0.400** | **0.412** | **0.452** | 14939 |
| 3 | **0.135** | **0.148** | **0.164** | _0.199_ | 6617 | 0.395 | 0.409 | 0.421 | 0.461 | 13581 |
| 4 | 0.137 | 0.150 | 0.166 | 0.201 | _5812_ | 0.400 | 0.412 | 0.424 | 0.466 | _11642_ |
| 5 | 0.142 | 0.155 | 0.170 | 0.204 | **5003** | 0.402 | 0.412 | 0.424 | 0.466 | **9915** |

**Table 13**
Results on Exchange with different input length

| Input length | Methods | Exchange (Univariate) | | | | Exchange (Multivariate) | | | |
| --- | --- | --- | --- | --- | --- | --- | --- | --- | --- |
| | | MSE | | MAE | | MSE | | MAE | |
| | | Mean | Std | Mean | Std | Mean | Std | Mean | Std |
| 96 | FDNet | **0.652** | *0.078* | **0.631** | *0.032* | **0.777** | **0.049** | **0.676** | **0.017** |
| | DLinear | *0.762* | **0.015** | *0.687* | **0.006** | 0.820 | *0.051* | 0.716 | *0.025* |
| | N-HiTS | 3.125 | 0.555 | 1.279 | 0.092 | *0.816* | 0.141 | *0.690* | 0.055 |
| 192 | FDNet | *0.911* | *0.155* | **0.751** | 0.092 | **1.084** | *0.090* | **0.783** | *0.030* |
| | DLinear | **0.910** | **0.027** | *0.749* | **0.012** | *1.020* | **0.050** | *0.800* | **0.022** |
| | N-HiTS | 3.520 | 0.198 | 1.535 | *0.047* | 1.287 | 0.321 | 0.850 | 0.092 |
| 288 | FDNet | **0.963** | *0.128* | **0.772** | 0.053 | **1.131** | *0.096* | **0.799** | *0.031* |
| | DLinear | 0.989 | **0.017** | *0.769* | **0.005** | *1.234* | **0.069** | *0.894* | **0.029** |
| | N-HiTS | 4.536 | 0.250 | 1.702 | *0.025* | 2.554 | 0.695 | 1.165 | 0.156 |
| 384 | FDNet | **0.863** | *0.106* | **0.729** | *0.045* | **1.163** | *0.102* | **0.808** | **0.034** |
| | DLinear | *1.133* | **0.025** | *0.818* | **0.012** | *1.472* | **0.099** | *0.992* | *0.035* |
| | N-HiTS | 5.216 | 0.576 | 1.908 | 0.148 | 3.597 | 0.852 | 1.412 | 0.176 |

### 6.10. More Experiments on Exchange Dataset

We discuss the forecasting accuracy variation of FDNet with longer input sequence length (> 96) on Exchange in this section. {N-HiTS, DLinear} are chosen as baselines as they have generally best performances in Table 5 under Exchange. DLinear takes the format of DLinear-I in virtue of its better forecasting performances on Exchange in its original paper. Input lengths are chosen among {96, 192, 288, 384}. Prediction length is set to 720. Forecasting results are shown in Table 13, including means and Stds of MSEs/MAEs during experiments of 10 times.

Compare prediction errors in Table 5/6 and Table 13, it could be observed that: (1) Both three methods behave worse if prolonging input sequence length under both univariate and multivariate forecasting. Causes of such phenomenon are pluralistic. We believe that there are at least two reasons. Firstly, Exchange is not a big dataset (Table 3) so that the reduction of training instances brought by longer input sequence will bring considerable impacts on training phases of all deep forecasting models. Additionally, Exchange is a dataset with the frequency of 1 day while others normally have frequency of 1 hour or even smaller. Therefore, time series within Exchange may not have such long-term dependency, e.g., when input sequence is 672, timespan of input sequence is nearly 2 years. (2) Though FDNet has relatively worse multivariate forecasting accuracy on Exchange when input sequence length is 96 in Table 5/6, it suffers slighter performance drop with longer input sequence length when compared with other baselines. Average errors of FDNet are mostly smaller than those of others under conditions when input lengths are among {192, 288, 384}. In several conditions, FDNet even achieves better forecasting accuracy with longer prediction length, illustrating superior LSTI handling capability of FDNet. (3) Forecasting performances of FDNet are also robust with longer input sequence lengths. Stds of FDNet and Dlinear are all very small and closed with each other while those of N-HiTS are relatively big.

In conclusion, it cannot be denied that FDNet will suffer performance drop while prolonging input sequence length (> 96) under Exchange, yet drop extents of FDNet are smaller than those of others, showing the superior LSTI handling capability and robustness of FDNet.

### 6.11. Analysis of Distribution Shift problem by KS Test

To examine whether extracting universal representations is possible for time series forecasting models under real-world time series forecasting tasks, we introduce Kolmogorov-Smirnov (KS) Test.

We perform KS test under six real-world time series benchmark datasets in Section 4. We randomly select 1000 sub-sequences of length 96 for each dataset and separately calculate Kolmogorov-Smirnov statistics, i.e., P-values, of the first selected sub-sequence and the rest. We take 0.05 as the margin P-value. Results in Table 2 has shown that universal representations for all local windows are formidable or even impossible to extract. To further examine this, universal representations for all local windows are formidable or even impossible to extract. To further examine this, we visualize distributions of two instances when experimenting with ETTm$_2$/ECL, which own highest/lowest reject rate, respectively as shown in Figure 10. Curves in Figure 10 clearly illustrate that different sub-sequences of both two datasets have different statistical properties or dynamics, which is identical to our aforementioned discovery that extracting universal representations is not necessary for time series forecasting models. Notice that even though the second instance of ECL owns P-value over the margin 0.05, distributions of these two curves have distinct differences.

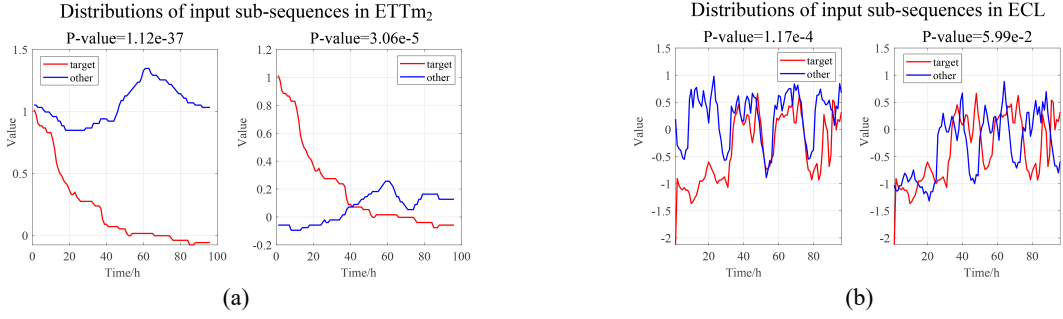

**Figure 10:** Distributions of two instances when experimenting with ETTm$_2$/ECL. Distributions of target and these two instances are quite different especially in ETTm$_2$.

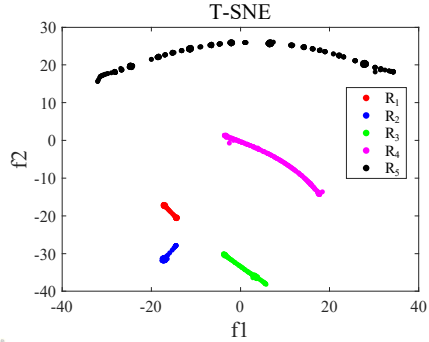

**Figure 11:** The visualization of final representations of elements in an input sequence by T-SNE under ETTm$_2$ univariate forecasting. The input sequence length is set to 672.

### 6.12. Visualization of Final Representation by T-SNE

FDNet with focal input sequence decomposition method have two expectations for final representations of each input sequence. The first one aims to independently extract representations of different input sub-sequences. It means that final representations of different input sub-sequences shall be distant from each other. Moreover, when the sub-sequence is closer to prediction sequence, the number of feature extraction layers increases so that the number of convolutional layers increases. It means that if the sub-sequence is closer to prediction sequence, final representations of its input elements shall be relatively closed to each other. To verify whether these are established in real-world applications, we visualize final representations of elements in an input sequence with length of 672 via T-SNE under ETTm$_2$ univariate forecasting as Figure 11 shows. This input sequence is randomly selected in the test dataset and it is divided into 5 sub-sequences by focal input sequence decomposition method. The prediction length is set to 96. Each data point in Figure 11 denotes the final representation of corresponding input element and its label $R_i$ refers to the $i$-th input sub-sequence it belongs to. The number of label increases with temporal distances between the input sequence it represents and prediction elements getting farther.

It is obvious from Figure 11 that data points with different colors are distant from each other. It illustrates that feature maps of different sub-sequences are independent with each other, which fits the first expectation. Meanwhile, data points of certain sub-sequence become more sparse as the sub-sequence gets longer and farther from prediction elements, which fits the second expectation. Therefore, two expectations of final representations of FDNet with focal input sequence decomposition method are all achieved, proving the rationality of its design.

# 7. Conclusion

In this paper, we propose FDNet whose core ideas contain decomposed forecasting formula and focal input sequence decomposition method. Built upon decomposed forecasting formula, FDNet is designed to only extract local fine-grained feature maps of input sequence, which is proved to be effective and feasible both theoretically and empirically. In addition, focal input sequence decomposition method solves long-standing LSTI problem by consecutively splitting and processing input sequence in a focal manner. Extensive experiments demonstrate that FDNet is simple but accurate, efficient, robust and practical for real-world time series forecasting.

# Acknowledgment

This work was partially supported by National Natural Science Foundation of China under grant #U19B2033, the National Key Research and Development Program of China under grant No. 2022YFB3904303 and the National Natural Science Foundation of China under grant No. 62076019.

## A. Kolmogorov-Smirnov Test

KS test is a nonparametric test to depict agreements between distributions of each two sequences. In essence, KS test describes the probability that they come from the same (but unknown) probability distribution. The Kolmogorov-Smirnov statistic measures a distance $D$ between the empirical distribution function of them and the value of $D$ is calculated as Equation 1.

$$D = \sup_{x} |F(\{x_i\}_{i=m}^{m+m_1}) - F(\{x_i\}_{i=n}^{n+n_1})| \tag{1}$$

Where $\{x_i\}_{i=m}^{m+m_1} / \{x_i\}_{i=n}^{n+n_1}$ refers to the sequence within timespan $[m, m+m_1]/[n, n+n_1]$, $F(\cdot)$ is the empirical distribution function and sup denotes the supremum function. For large samples, the null hypothesis is rejected at level $\alpha$ if the calculated value $D$ satisfies Inequality 2 so that P-value is smaller than level $\alpha$ as Equation 3 shows. It means that if P-value is small, the null hypothesis is more likely to be rejected, i.e., these two distributions are more likely to be different.

$$D > \sqrt{-\frac{1}{2}\ln\frac{\alpha}{2}} \times \sqrt{\frac{m_1 + n_1}{m_1 \cdot n_1}} \tag{2}$$

$$\text{P} - \text{value} = 2e^{-2D^2 \frac{m_1 \cdot n_1}{m_1 + n_1}} < \alpha \tag{3}$$

It can be deduced that if there exist universal properties of input sequences, statistical properties or dynamics of local input sequences will be similar to each other so that their P-values will be bigger compared with the margin P-value.

## B. T-distributed Stochastic Neighbourhood Embedding

T-distributed Stochastic Neighbourhood Embedding (T-SNE) [65] is a statistical method to visualize high-dimensional data by giving each datapoint a location in a two-/three-dimensional map by minimizing KL divergence between a joint probability distribution in the high-dimensional space and a joint probability distribution in the low-dimensional space. It is built upon Stochastic Neighbor Embedding (SNE) [66] and replaces a Gaussian distribution with a Student-t distribution to better compute similarity between each two points in the low-dimensional space. Specifically, similar high-dimensional datapoints are modeled by nearby points in a two-/three-dimensional map and dissimilar ones are modeled by distant points in a two-/three-dimensional map with high probability.

## C. Supplimentary Experiment

### C.1. Detailed Explanation for abandoning ILI dataset

**ILI** (Influenza-like Illness)[3] dataset, contains weekly recorded influenza-like illness patients data obtained by Centers for Disease Control and Prevention of the United States between 2002 and 2021. The size of ILI is 966, its dimension is 8 and its frequency is 7 days as Table 14 shows.

As Table 3/14 shows, the size of ILI is dozens times smaller than those of other six benchmark datasets. 672 is selected as input sequence length to examine LSTI problem handling capabilities of models in this paper. Considering that train/val/test is 70%/10%/20%, ILI dataset can not satisfy the input sequence length of 672 and abundant instances. Moreover, there are only hundreds of instances available for training even though input sequence length is shortened.

**Table 14**
Details of ILI

| Dataset | Size | Dimension | Frequency |
|---------|------|-----------|-----------|
| ILI     | 966  | 8         | 7 days    |

---

[3] ILI dataset was acquired at: https://drive.google.com/drive/folders/1ZOYpTUa82_jCcxIdTmyr0LXQfvaM9vIy?usp=sharing

**Table 15**
Details of M4

| Sub-dataset | Input length ($L_{in}$) | Prediction length ($L_{out}$) | Periodicity ($m$) | Frequency | Number of instances |
|---|---|---|---|---|---|
| M4-Yearly | 12 | 6 | 1 | 1 year | 23000 |
| M4-Quarterly | 24 | 8 | 4 | 1 quarter | 24000 |
| M4--Monthly | 72 | 18 | 12 | 1 month | 48000 |
| M4-Weekly | 65 | 13 | 1 | 1 week | 359 |
| M4-Daily | 84 | 14 | 1 | 1 day | 4227 |
| M4-Hourly | 336 | 48 | 24 | 1 hour | 414 |
| Total | | | | | 100000 |

## C.2. Supplementary Experiment on M4 Dataset

To persuasively show that FDNet excels in univariate forecasting, we supplement experiments results on six sub-datasets of M4 [64]. Their introductions are shown below and numeral statistics are shown in Table 15.

*M4* dataset [64], a competition dataset specialized for univariate forecasting, contains 100K real-world time series instances categorized by sampling frequencies {1 year, 1 quarter, 1 month, 1 week, 1 day, 1 hour}. {M4-Yearly, M4-Quarterly, M4-Monthly, M4-Weekly, M4-Daily, M4-Hourly} are sub-datasets of M4 with corresponding sampling frequencies. The ratio of train/val is 3/1.

The settings are referenced from NBEATS [46]. The input/prediction lengths are {12, 24, 72, 64, 84, 336}/{6, 8, 18, 13, 14, 48} for corresponding sub-datasets {M4-Yearly, M4-Quarterly, M4-Monthly, M4-Weekly, M4-Daily, M4-Hourly}. The evaluation metrics are SMAPE (Symmetric Mean Absolute Percentage Error, Equation 4) and OWA (Overall Weighted Average, Equation 5) following M4 competition [64]. naïve2 is the seasonally-adjusted forecast model provided by [64] and $m$ is the periodicity of series. The training strategy for every baseline is training the same model for each sub-dataset. In others words, we do not train 100K models separately for each instance but train six models separately for all instances in six sub-datasets with each baseline. This strategy is referenced from https://github.com/ServiceNow/N-BEATS, an implementation of NBEATS on M4. Though the results of all baselines may be not better than the results of competitive participants in M4 competition, whose parameters and settings are well-adjusted, they are persuasive to compare the univariate forecasting capabilities of all baselines as these baselines are evaluated under the same setting and common hyper-parameters.

$$\text{SMAPE} = \frac{200}{T - t_0} \sum_{t \in [t_0+1,T]} \frac{|x_t - \hat{x}_t|}{|x_t| + |\hat{x}_t|} \tag{4}$$

$$\text{MASE} = \frac{1}{T - t_0} \sum_{t \in [t_0+1,T]} \frac{|x_t - \hat{x}_t|}{\frac{1}{T-m} \sum_{j \in [m+1,T]} |x_j - x_{j-m}|}$$

$$\text{OWA} = \frac{1}{2} \left( \frac{\text{SMAPE}}{\text{SMAPE}_{\text{naïve2}}} + \frac{\text{MASE}}{\text{MASE}_{\text{naïve2}}} \right) \tag{5}$$

The forecasting results on six sub-datasets of M4 are shown in Table 16. When compared with FEDformer/ETSformer/Triformer/Non-stationary Transformer/PatchTST/SCINet/DLinear/N-HiTS/FiLM/Linear/AutoAI-TS/CoST, FDNet yields 26.1%/36.5%/86.3%/39.4%/42.9%/19.6%/64.1%/19.6%/45.9%/64.4%/45.6%/72.8% relative SMAPE reduction and 26.1%/32.2%/89.7%/40.3%/40.4%/18.9%/72.8%/17.7%/46.5%/69.6%/45.8%/73.9% relative OWA reduction in general. This illustrates that compared with other deep forecasting models, FDNet still has an edge on univariate forecasting when handling more complicated and formal univariate forecasting conditions.

**Table 16**
Results of univariate forecasting under M4 dataset

| Methods | Metrics | M4-Yearly | M4-Quarterly | M4-Monthly | M4-Weekly | M4-Daily | M4-Hourly |
|---|---|---|---|---|---|---|---|
| FDNet | SMAPE | **11.282** | **9.915** | 9.773 | **9.292** | **3.321** | **15.684** |
| | OWA | **0.846** | **0.954** | 0.896 | **1.257** | **1.140** | **1.329** |
| FEDformer | SMAPE | 16.479 | 12.293 | 16.064 | 12.816 | 4.449 | 18.085 |
| | OWA | 1.235 | 1.183 | 1.472 | 1.734 | 1.528 | 1.532 |
| ETSformer | SMAPE | 17.364 | 14.270 | 15.638 | _10.394_ | 3.634 | 32.045 |
| | OWA | 1.302 | 1.373 | 1.433 | _1.406_ | 1.248 | 2.714 |
| Triformer | SMAPE | 65.685 | 67.819 | 65.293 | 86.161 | 79.622 | 68.425 |
| | OWA | 4.924 | 6.525 | 5.984 | 11.657 | 27.337 | 5.796 |
| Non-stationary Transformer | SMAPE | 27.878 | 17.524 | 15.923 | 12.036 | 7.029 | _17.489_ |
| | OWA | 2.090 | 1.686 | 1.459 | 1.628 | 2.413 | _1.481_ |
| PatchTST | SMAPE | 14.340 | 11.999 | 10.400 | 14.923 | 4.450 | 47.657 |
| | OWA | 1.075 | 1.155 | 0.953 | 2.019 | 1.528 | 4.037 |
| SCINet | SMAPE | _13.716_ | _10.777_ | 9.215 | 14.139 | _3.492_ | 22.376 |
| | OWA | _1.028_ | _1.037_ | **0.845** | 1.913 | _1.199_ | 1.895 |
| DLinear | SMAPE | 26.703 | 10.862 | 12.726 | 49.301 | 27.680 | 38.037 |
| | OWA | 2.002 | 1.045 | 1.166 | 6.670 | 9.504 | 3.222 |
| N-HiTS | SMAPE | 15.181 | 12.096 | 9.814 | 11.643 | 3.518 | 21.426 |
| | OWA | 1.138 | 1.164 | 0.900 | 1.575 | 1.208 | 1.815 |
| FiLM | SMAPE | 19.903 | 18.281 | 17.291 | 14.224 | 7.192 | 32.686 |
| | OWA | 1.492 | 1.759 | 1.585 | 1.924 | 2.469 | 2.769 |
| Linear | SMAPE | 48.780 | 11.367 | _9.733_ | 50.397 | 18.251 | 27.999 |
| | OWA | 3.656 | 1.094 | _0.892_ | 6.818 | 6.266 | 2.372 |
| AutoAI-TS | SMAPE | 19.208 | 15.617 | 13.923 | 10.981 | 7.672 | 41.536 |
| | OWA | 1.440 | 1.503 | 1.276 | 1.486 | 2.634 | 3.518 |
| CoST | SMAPE | 16.302 | 17.991 | 119.275 | 15.036 | 17.496 | 31.476 |
| | OWA | 1.222 | 1.731 | 10.932 | 2.034 | 6.007 | 2.666 |

## D. An overview of FDNet Architecture in Details

An overview of FDNet/FUNet architecture in details is shown in Table 17/19. 'DFE-ICOM'/'DFE-initial' refers to decomposed feature extraction layer with/without ICOM and their detailed architectures are shown in Table 18/20 separately.

**Table 17**
An overview of FDNet architecture in details

| Output sequence | | | | |
|---|---|---|---|---|
| Add | | | | |
| Linear$_5$ | Linear$_4$ | Linear$_3$ | Linear$_2$ | Linear$_1$ |
| Flatten$_5$ | Flatten$_4$ | Flatten$_3$ | Flatten$_2$ | Flatten$_1$ |
| DFE-initial×1 | DFE-initial×2 | DFE-initial×3 | DFE-initial×4 | DFE-initial×5 |
| Embedding$_5$ | Embedding$_4$ | Embedding$_3$ | Embedding$_2$ | Embedding$_1$ |
| Input$_5$ | Input$_4$ | Input$_3$ | Input$_2$ | Input$_1$ |
| 1/2 | 1/4 | 1/8 | 1/16 | 1/16 |
| Split | | | | |
| Input sequence ($\{x_{i,1:t_0}\}_{i=1}^{N}$) | | | | |

**Table 18**
DFE-initial components in details

| Input feature map |
|---|
| $1 \times 1$ Conv |
| WN, Dropout ($p = 0.1$), Gelu |
| $3 \times 1$ Conv padding = (1,0) |
| Add, WN, Dropout ($p = 0.1$), Gelu |
| $1 \times 1$ Conv |
| WN, Dropout ($p = 0.1$), Gelu |
| $3 \times 1$ Conv padding = (1,0) |
| Add, WN, Dropout ($p = 0.1$), Gelu |
| Output feature map |

**Table 19**
An overview of FUNet architecture in details

| | | Output sequence | | |
|---|---|---|---|---|
| | | Add | | |
| Linear$_5$ | Linear$_4$ | Linear$_3$ | Linear$_2$ | Linear$_1$ |
| Flatten$_5$ | Flatten$_4$ | Flatten$_3$ | Flatten$_2$ | Flatten$_1$ |
| DFE-ICOM×4 | DFE-ICOM×3 | DFE-ICOM×2 | DFE-ICOM×1 | DFE-ICOM×1 |
| Embedding$_5$ | Embedding$_4$ | Embedding$_3$ | Embedding$_2$ | Embedding$_1$ |
| Input$_5$ | Input$_4$ | Input$_3$ | Input$_2$ | Input$_1$ |
| 1/2 | 1/4 | 1/8 | 1/16 | 1/16 |
| | | Split | | |
| | | Input sequence ($\{x_{i,1:t_0}\}_{i=1}^N$) | | |

**Table 20**
DFE-ICOM components in details

| Input feature map | |
|---|---|
| Multi-head full attention ($d=8, h=N$) | |
| $1 \times 1$ Conv | |
| Add, WN, Dropout ($p=0.1$), Gelu | |
| $3 \times 1$ Conv stride = (2,1), padding = (1,0) | |
| WN, Dropout ($p=0.1$), Gelu | $3 \times 1$ Maxpooling stride = (2,1), padding = (1,0) |
| $3 \times 1$ Conv padding=(1,0) | |
| WN, Dropout ($p=0.1$), Gelu | |
| Add | |
| Output feature map | |

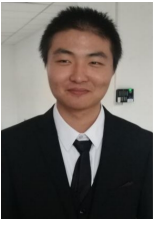

**Li Shen** received his B.S. degree in Navigation and Control from Beihang University in China at the School of Automation Science and Electrical Engineering. He is currently completing a Doctor degree in Electronic Information Engineering at the Beihang Institute of Unmanned System. His research interests include Computer Vision and Time Series Forecasting.

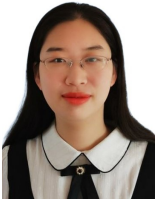

**Yuning Wei** received her B.S. degree in Agricultural Electrification from Nanjing Agricultural University in China at the School of Engineering. She is currently completing a MA.Eng in Electronic Information Engineering at the Beihang Institute of Unmanned System. Her research interests include Deep Learning and Time Series Forecasting.

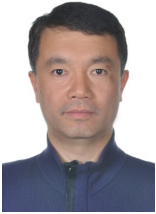

**Yangzhu Wang** received his B.S, MA.Eng and Ph.D degrees in Measurement and Control Technology and Instrumentation Program from Beihang University in China at the Beihang Institute of Unmanned System. He is currently a researcher fellow of the Flying College of Beihang University and Beihang Institute of Unmanned System. His research interests include Computer Vision, Unmanned System, and Measurement and Control Technology.

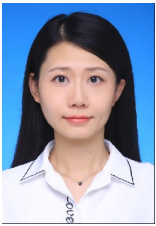

**Huaxin Qiu** received the B.S. degree in automation and the Ph.D. degree in guidance, navigation, and control from Beihang University, Beijing, China, in 2014 and 2019, respectively. From 2019 to 2022, she was a Postdoctoral Researcher with the Qian Xuesen Laboratory of Space Technology, China Academy of Space Technology, Beijing. Since 2022, she has been an Associate Professor with the Flying College, Beihang University. Her main research interests include swarm intelligence with its application to distributed decision-making and optimization in multiagent systems.